\documentclass[a4paper,11pt]{article}
\advance\hoffset by -0.5truein
\usepackage[utf8]{inputenc}
\usepackage{graphicx}
\usepackage{epstopdf}
\usepackage{diagbox}
\usepackage{tikz}
\usepackage{booktabs}
\usepackage{graphicx}
\usepackage{booktabs}
\usepackage{dcolumn}

\usepackage[colorinlistoftodos]{todonotes}
\usepackage[normalem]{ulem}

\usepackage{amsmath}
\usepackage{amssymb}
\usepackage{inputenc}
\usepackage{xcolor}
\usepackage{url}
\usepackage{pdflscape}
\usepackage{amssymb,latexsym}

\usepackage{enumerate}

 \usepackage[T1]{fontenc}

\usepackage{subcaption}

\newfont{\mm}{eufm10 scaled 1200}
\begin{document}


\centerline{\Large\bf Robust Optimization of Spring Design}
\vspace{0.1cm}
\centerline{\Large\bf under Variable Manufacturing Tolerances}

\vspace{0.55cm}

\centerline{\large Grzegorz Sroka } \vspace{4mm}
\centerline{\footnotesize\it Department of Analysis Nonlinear, Rzesz\'ow University of Technology,}
\centerline{\footnotesize\it Powsta\'nc\'ow Warszawy 12, 35-959 Rzesz\'ow, Poland}
\centerline{\footnotesize\it  e-mail: gsroka@prz.edu.pl}
\vspace{0,4cm}
\centerline{\large Sławomir T. Wierzchoń} \vspace{4mm}
\centerline{\footnotesize\it   Institute of Computer Science, Polish Academy of Sciences,}
\centerline{\footnotesize\it   ul. Jana Kazimierza 5, 01-248 Warsaw, Poland}

\vspace{0.35cm}


\vspace{0.3cm}

{\footnotesize  {\bf Abstract:} A reliable assessment of an optimizer’s robustness requires tracking controlled changes in parameters to identify sources of algorithmic weakness. Such an approach was applied to the problem of designing tension/compression springs with three continuous variables describing the spring geometry. 

Three additional independent binary parameters reflect material variability, manufacturing-related geometric deviations, and an additional safety margin. Without expanding the decision space, these parameters can be combined in eight ways to generate variants that allow for direct comparison.

In this study we distinguish between robustness effects and the dimension of the problem, which allows tracking changes in feasibility while keeping the nominal mass objective constant. Seven optimization methods and eight hybridized their variants are used for the analysis. The hybridization mechanism applies the operator selected by the prediction procedure presented in Section 3. All methods are evaluated under the same experimental conditions. Each method is run thirty times for each variant, with a maximum of 30,000 function evaluations permitted per run.

The final quality assessment takes into account the geometry of the feasibility region, search trajectories, performance variations across landscapes and the computational cost of fixed objectives. Together, these factors allow us to determine the quality of the solution, reproducibility, feasibility control, and the effectiveness of optimization.

An interesting observation was that hybridization can mitigate cost or feasibility losses associated with a specific configuration. While it compensates for a specific weakness in the underlying mechanism, it does not provide universal resilience.}

\vspace{0.15cm}


\vspace{0.15cm}

\section{Introduction}

The classical Tension/Compression Spring Design (TCSD) problem adopts a deliberately compact formulation. 
The \texttt{3-variable structure} of the model provides an engineering framework for the comparison of optimization 
methods under common and reproducible conditions. Cadet and Paredes in paper \cite{GC2024} explicitly describe this role, 
stating that its purpose: 
\begin{center}
\textbf{``is not to be realistic but to facilitate the comparison of hundreds of optimization
algorithms by using similar initial conditions''.}
\end{center}
The present study is primarily motivated by this benchmark role.

Nevertheless, this simplicity creates a methodological limitation. A single nominal TCSD instance represents only one feasibility structure. It therefore cannot show whether the observed algorithm behaviour persists as robustness related conditions change systematically. Increasing model complexity would address a different question. It would introduce additional variables, objectives, or physical dependencies. This creates a gap between nominal TCSD benchmarking and richer robust spring design models. A controlled framework is therefore required. It should allow robustness-related conditions to vary while the underlying optimisation problem remains unchanged.

The aim of this study is to determine how such controlled changes affect algorithm behaviour, algorithm differentiation, and the response to a common hybridisation mechanism. To this end, we introduce the 
$MPR$ parameterisation (e.g., Section \ref{subsec:MPR}). The classical decision vector and nominal mass objective are preserved. Three binary configuration parameters, $M$, $P$, and $R$, independently activate: \textbf{\textsc{material variability, manufacturing related geometric perturbations, and a normalised requirements margin}}. These parameters define the problem instance and are not optimisation variables. Their full $2^3$ combination yields eight matched configurations (e.g., Table 
\ref{tab:MPR_configs}.), ranging from nominal TCSD to the case in which all three mechanisms are active. The search dimension remains three. Only the feasibility conditions change, in a controlled and engineering interpretable manner.

MPR transfers the logic of independently controlled benchmark configurations from numerical test functions to a constrained engineering design problem. It is not intended to replace detailed industrial spring sizing or probabilistic reliability models. Instead, it provides a compact diagnostic framework between a single nominal benchmark and richer robust design formulations. Within the same underlying design model, both individual and combined robustness mechanisms can be examined. This facilitates the interpretation of changes in attainable mass and algorithm behaviour on a common basis.

The present experimental study evaluates eight baseline algorithms with distinct search mechanisms (e.g., Section \ref{RW}, Table \ref{tab:MPR_all_results}). It also encompasses seven predicted candidate hybrid variants tested under the same protocol. The same hybridisation principle is applied to all compatible baseline methods. This enables direct comparison of responses to the MPR configurations across different search mechanisms. It also allows the hybridisation principle to be examined as the feasibility structure changes.

The analysis extends beyond final objective values. The objective constraint violation geometry is examined. This reveals how $MPR$ reshapes the feasible structure. A performance space approach is adapted to constrained single objective optimisation. It characterises algorithm differentiation throughout the search process. Its constraint component is aligned with \texttt{the quadratic worst-case violation measure adopted for $MPR$} (e.g., Eq.~\eqref{eq:Viol}). Pairwise separation is assessed from runtime behaviour. A complementary fixed-target analysis is also performed. It quantifies the computational effort required to attain prescribed feasible objective accuracies.

The paper is divided into eight sections. The research problem and objectives are introduced in Section 1, followed by related work in Section 2. In Section 3, the predicted-candidate operator is presented. The classical TCSD and its $MPR$ extension are formulated in Section 4. The experimental protocol is detailed in Section 5, and the results are presented in Section 6. In Sections 7 and 8, the findings are discussed and the conclusions presented, respectively.

\section{Related Work}
\label{RW}

Benchmarking practices for the TCSD were examined by Tzanetos and Blondin in \cite{AT2023} across 60 metaheuristics. Following validation and comparability checks, only 18 remained in the final ranking. The nominal \texttt{TCSD} (for 3-variable problem) has also been used by: Alawadi et al. in \cite{SA2026}, Sroka and Wierzchoń in  \cite{GS2026}, Wu et al. in \cite{DW2025}, Alsayyed et al. in \cite{OA2023}, {\DJ}ur{\dh}ev in \cite{MD2021}, Oszut et al. in \cite{MO2021},  Faramarzi et al. in \cite{AF2020}, Mirjalili and Lewis in \cite{SM2016} and Mirjalili et al. in~\cite{SM2014} to validate new optimisation methods. Its continued use reinforces its status as an engineering benchmark. This role is further supported by Kumar et al. in \cite{AK2020}, who included it in a suite of 57 real-world constrained problems. The 3-variable spring problem is designated as \texttt{RC17} (i.e., problem 17 of the \textbf{CEC 2020} Competitions on Evolutionary Computation benchmark suite). The optimisation of spring design can be affected by constraint treatment. The numerical difficulties analysed by Rodriguez et al. in \cite{ER2004} required problem-specific constraint-handling strategies.

Within the broader \texttt{Constrained Single-Objective Optimization Problem} \texttt{(CSOP)} context, Hellwig and Beyer in \cite{MH2019} reviewed benchmark design and constraint characteristics. More broadly, Jin and Branke in \cite{MH2005} emphasise the effect of uncertainty in design variables and environmental parameters on solution robustness.

Against this broader backdrop, spring optimisation can also support the practical synthesis of designs. 
In this context, the framework proposed by Paredes et al. in \cite{MP2002} is based on interval-valued specifications. Its constraints support early-stage dimensional synthesis under design, standard, and manufacturing limits. In turn, Trabelsi et al. in \cite{HT2013}, \cite{HT2015} combined interval computation with a \texttt{Constraint Satisfaction Problem (CSP)} formulation coupling static and dynamic sizing requirements. The approach aimed to identify feasible design variable intervals from the outset. Regarding constraint handling, Kulkarni et al. in \cite{OK2018} evaluated static and dynamic penalties for the \texttt{TCSD} problem using the \texttt{Cohort Intelligence} metaheuristic (i.e., based on interactive learning). From a robustness standpoint, Silva et al. in \cite{ADGS2015} examined a fuzzy formulation with $\pm5\%$ uncertainty in material density, applied load, and allowable shear stress. However, they found that the extent of robustness gains depended on the problem (for 3-variable spring case), while the method could require many cost function evaluations. More recently, Cadet and Paredes in \cite{GC2024} incorporated manufacturing uncertainty and material variability into robust spring sizing, targeting low mass and a high probability of constraint satisfaction. In their subsequent study in \cite{GC2026}, Cadet and Paredes improved reliability through geometry optimisation and machine tuning for the softest and stiffest wire cases. While these studies focus on robust and reliable spring design, the present work uses matched benchmarks to analyse optimiser behaviour.

More broadly, the parameterised benchmarking approach proposed by Opara et al. in \cite{KO2020} informed the design of the present benchmark. They systematically varied transformations used in the \textbf{CEC 2017} benchmark (e.g., Awad et al. in \cite{NHA2016}), including: \texttt{rotation, scaling and translation.} This produced function variants with interpretable characteristics. In a related development, Mohamed et al. in \cite{AWM2020} used binary control of \texttt{bias, shift and rotation} in the \textbf{CEC 2021} benchmark. A subsequent evaluation by Mohamed et al. in \cite{AWM2023} showed that metaheuristic performance varied across these configurations. A parameterised \textbf{CEC 2017}
(e.g., Awad et al. in \cite{NHA2016}) implementation was developed by Sroka and Wierzchoń in \cite{GS2026a} with independent control of the three benchmark transformations. It was used to diagnose the hybrid Marine Predators Algorithm (hMPA, where $h$ denotes the hybrid variant. See, Section \ref{sec:predicted_candidate_operator}).
Statistically significant differences were found among configurations at every tested dimension (for $dim\in\{10, 30, 50, 100\}$). However, no consistent ordering emerged across functions. The predicted candidate operator, introduced by Oszust et al. in \cite{MO2021}, links this diagnostic line to algorithmic hybridisation. 

Our experimental study includes eight base algorithms and seven hybrid variants. Their selection was driven primarily by differences in search mechanisms. Covariance Matrix Adaptation Evolution Strategy (CMA-ES, e.g., Hansen et al. in \cite{NH2001},\cite{NH2003},\cite{NH2004}) adapts a Gaussian search distribution through its mean, step size, and covariance matrix. Differential Evolution (DE, e.g., Storn and Price in \cite{RS1997}) generates trial solutions using differential mutation and crossover. As Tanabe and Fukunaga describe in \cite{RT2014}, Success-History Based Adaptive DE with LPSR (L-SHADE) combines success history based parameter adaptation with linear population size reduction. In turn,                   
Marine Predators Algorithm (MPA), introduced by Faramarzi et al. in \cite{AF2020}, employs a phase dependent search based on Brownian and L\'evy movements. In swarm search, Standard Particle Swarm Optimisation (SPSO2011, e.g., Clerc in \cite{MC2012}) uses personal and neighbourhood bests within an adaptive random topology. Whale Optimization Algorithm (WOA, e.g., Mirjalili and Lewis in \cite{SM2016}) uses encircling and spiral search for exploitation, together with randomised exploration. In the work of Faramarzi et al. in \cite{AFeo2020}, the Equilibrium Optimizer (EO) updates each particle using a randomly selected member of the equilibrium pool. Within evolutionary search, Genetic Algoritm (GA, e.g., Deb and Kumar in \cite{KD1995} and Deb in \cite{KD2000}) relies on selection, crossover, and mutation.

The potential of hybridisation has also been explored for several of these search mechanisms. Such examples include:           
L-SHADE-CMA-ES by Mohamed et al. in \cite{AWM2017}, GA-CMA-ES by  Guo, Hou and He in \cite{ZG2024}, EO-DE by Moharam et al. in \cite{AM2024}, WOA-MPA by Li and Xu in \cite{YL2026} and WOA-DE by Zhongyu in \cite{WZ2019}. Mechanistically related approaches include the MPA-PSO, GA-PSO, CMA-ES-PSO, and  SHADE–WOA hybrids proposed by: Shemshaki et al. in \cite{ES2025}, Juang in \cite{CFJ2004}, Xu et al. in \cite{PX2019}, and Chakraborty et al. in \cite{SC2021}, respectively.

This diversity is integral to the experimental design. It allows us to systematically analyse the responses of algorithms to controlled $MPR$ configurations (e.g., Section \ref{subsec:MPR}). These responses are then compared across different search mechanisms. The same \texttt{predicted candidate operator} (e.g., Section \ref{sec:predicted_candidate_operator}) is applied to all compatible methods. This provides a consistent basis for comparing hybrid responses across different host mechanisms.

Introduced by Hansen et al. in \cite{NH2001},\cite{NH2003},\cite{NH2004}, CMA-ES requires separate treatment.
The predicted candidate operator replaces selected population members with externally generated candidates. Oszust et al. in \cite{MO2021} show that this replacement process can interact with internal population dependencies. 
In CMA-ES, however, the ranking of sampled offspring directly influences the adaptation of the Gaussian search model.
Therefore, such external replacement could disrupt the coupling between sampling and model adaptation. The operator cannot be incorporated into standard CMA-ES through direct population replacement without redesigning its update mechanism. Consequently, the standard form of \textbf{CMA-ES is retained and no hCMA-ES variant is considered}.
\vspace{0.2cm}

This distinction defines the scope of hybridisation considered below.
\section{Predicted candidate operator}
\label{sec:predicted_candidate_operator}

This study considers hybrid variants that incorporate the predicted candidate
operator introduced by Oszust, Sroka, and Cymerys in \cite{MO2021} into the
population update cycle of the host metaheuristic. This operator introduces a
small number of predicted solutions derived from information available in the
current population. The native search equations of the host method remain
unchanged.
Let:
$$X=\{x^{(1)},\ldots,x^{(n)}\},
\qquad
Y=\{F(x^{(1)}),\ldots,F(x^{(n)})\},$$
where $n$ denotes the population size,
$x^{(j)}\in\mathbb{R}^{dim}$ ($\mathbb{R}$ denoting the set of real
numbers and $dim$ the problem dimension), and $F(x^{(j)})$ is the scalar
fitness value of the $j$-th individual ($j=1,\ldots,n$).

The population is evaluated before the operator is
applied. Its members can therefore be ordered directly according to the
values contained in $Y$.

For each predicted candidate, Oszust et al. in~\cite{MO2021} define the target fitness level as follows:
$\widetilde F=\operatorname{rand}(a,b)\cdot
\frac{1}{n}\cdot\sum_{j=1}^{n}F(x^{(j)}).$
Here, $\operatorname{rand}(a,b)$ denotes a value drawn from the uniform
distribution on the open interval $(a,b)$. This interval depends on the
current stage of the optimisation process. During the first half of the run,
it is $(0.5,1.5)$; during the next quarter, it is $(0.95,1.05)$; and during
the final quarter, it is $(0.99,1.01)$.

The decision variables of the new candidate are obtained through inverse
regression calibration. For each coordinate $k=1,\ldots,dim$, the current
population provides the pairs
$(x_k^{(j)},F(x^{(j)}))$, $j=1,\ldots,n$. In the linear variant used here, a least-squares relation is estimated
separately for each coordinate:
$$F(x^{(j)})
\approx
\widehat{\beta}_{0,k}+
\widehat{\beta}_{1,k}x_k^{(j)},
\quad j=1,\ldots,n \quad \text{and} \quad k=1,\ldots,dim,$$
where $\widehat{\beta}_{0,k}$ and $\widehat{\beta}_{1,k}$ are the
least-squares estimates of the regression coefficients for the $k$-th
coordinate. The coordinate associated with the target level is then obtained as follows:
$$\widetilde x_k=\frac{\widetilde F-\widehat{\beta}_{0,k}}
     {\widehat{\beta}_{1,k}},
\qquad k=1,\ldots,dim.$$
The coordinate-wise estimates are then combined to produce the predicted candidate:
$\widetilde x=(\widetilde x_1,\ldots,\widetilde x_{dim})$.

The discussion below focuses on the operator's replacement setting,
computational complexity, and prior hybrid applications.

\paragraph{{\large\textbullet}\ \textsc{Predicted-candidate replacement rate.}}

In a minimisation problem, the individuals with the highest fitness values
are selected for replacement. The number of replaced individuals, $g$, is defined as follows:
$$g=\left\lfloor \texttt{newNopar}\cdot n \right\rfloor,$$
where $\texttt{newNopar}$ denotes the fraction of the current population replaced by the predicted candidates, $n$ as above, and $\lfloor\cdot\rfloor$ stands for the floor function.

The sensitivity analysis by Oszust et al.~\cite{MO2021} examined
replacement rates from $10\%$ to $90\%$ of the population in $10\%$ increments.
Average performance remained relatively stable over the $10-60\%$ range,
while rates up to $30\%$ were reported as generally safe across the tested
metaheuristics. Nevertheless, the effect of $g$ depended on the host
metaheuristic, indicating that individual tuning could further improve
hybrid performance. Their principal comparisons used a common $10\%$
replacement rate across the hybrid methods. Consistent with this protocol,
we set $\texttt{newNopar}=0.1$ for every hybrid algorithm, without
host-specific tuning. This keeps the replacement intensity constant across the 
baseline hybrid comparisons for the parameterised spring problem.

\paragraph{{\large\textbullet}\ \textsc{Computational overhead of the hybrid operator.}}

The hybridisation step involves additional population processing
operations. It does not, however, increase the objective-evaluation count.
According to Oszust et al.~\cite{MO2021}, population sorting requires
$O(n\cdot \log(n))$ operations. Target-value construction requires $O(g\cdot n)$.
Candidate reconstruction requires $O(g\cdot dim \cdot n\cdot p)$, while
replacement requires $O(g\cdot dim)$. The linear calibration 
model adopted in this study uses $p=1$. The additional asymptotic 
complexity of the operator is as follows:
$O\!\left(n\cdot (\log(n) + g) + g\cdot dim \cdot (n + 1)\right).$

\paragraph{{\large\textbullet}\ \textsc{Metaheuristics hybridised with the operator.}}

In the original study by Oszust et al. in~\cite{MO2021}, the operator was
integrated with ten population based metaheuristics:
Particle Swarm Optimization
(PSO) by Aydilek in~\cite{IBA2018}, Marine Predators Algorithm (MPA) by
Faramarzi et al. in~\cite{AF2020}, Salp Swarm Algorithm (SSA) by Mirjalili
in~\cite{SM2017}, Butterfly Optimization Algorithm (BOA) by Arora and Singh
in~\cite{SA2019}, and Grasshopper Optimization Algorithm (GOA) by Saremi,
Mirjalili and Lewis in~\cite{SS2017}. The remaining algorithms were:
Dragonfly Algorithm (DA) by Mirjalili in~\cite{SM2016a}, Moth-Flame
Optimization (MFO) by Mirjalili in~\cite{SM2015}, Grey Wolf Optimizer
(GWO) by Mirjalili et al. in~\cite{SM2014}, Multi-Verse Optimizer (MVO)
by Mirjalili et al. in~\cite{SM2016b}, and Wind Driven Optimization (WDO)
by Bayraktar in~\cite{ZB2013}.

The operator was later applied by Sroka and Wierzchoń in~\cite{GS2026}
to additional metaheuristics. These included:
Standard Particle Swarm Optimisation
(SPSO2011) by Clerc in~\cite{MC2012}, Hunger Games Search (HGS) by Yang
et al. in~\cite{YY2021}, Slime Mould Algorithm (SMA) by Li et al.
in~\cite{SL2020}, and Attraction Repulsion Optimization Algorithm (AROA)
by Cymerys and Oszust in~\cite{KC2024}. The extended set also includes:
Success-History-based Adaptive Differential Evolution (SHADE) by Tanabe
and Fukunaga in~\cite{RT2013}, Improved Multi-Operator Differential
Evolution (IMODE) by Sallam et al. in~\cite{KMS2020}, Harris Hawks
Optimization (HHO) by Heidari et al. in~\cite{AAH2019}, and Double
Mutational Salp Swarm Algorithm (DMSSA) by Lin et al. in~\cite{CLin2023}.
These prior applications provide the methodological basis for 
the algorithm set examined in this study.

\section{Helical compression spring optimization problem}
\label{sec:spring}

\subsection{Classical TCSD formulation}
\label{subsec:classical_TCSD}

The problem considered in this study, as studied by Belegundu and Arora in \cite{ADB1985}, is the classical TCSD. It is widely used as a constrained benchmark in mechanical engineering and metaheuristic optimisation. The objective is to minimise spring material usage while meeting mechanical and geometric requirements.
The spring geometry is described by three continuous decision variables:
the wire diameter $x_1$, the mean coil diameter $x_2$, and the number
of active coils $x_3$ (with a continuous relaxation adopted for $x_3$).
The standard benchmark bounds are:
\begin{equation}
0.05 \le x_1 \le 2, \qquad 
0.25 \le x_2 \le 1.30, \qquad
2 \le x_3 \le 15\
\label{eq:DzOmega}
\end{equation}
For a candidate design, the objective function is proportional to the spring material's volume and, for a constant density of the material, to its mass:
\begin{equation}
f(x)=(x_3+2)x_2x_1^2
\label{eq:classical_spring_objective}
\end{equation} where $x = (x_1,x_2,x_3)$.
The design must satisfy the following four standard mechanical and geometric requirements:
\begin{equation}
\label{eq:classical_constraints}
\begin{alignedat}{2}
g_1(x) &=
1-\frac{x_2^3x_3}{71785\,x_1^4}
\le 0,
\qquad&
g_2(x) &=
\frac{4x_2^2-x_1x_2}
{12566\,(x_2x_1^3-x_1^4)}
+\frac{1}{5108\,x_1^2}-1
\le 0,
\\[0.8ex]
g_3(x) &=
1-\frac{140.45\,x_1}{x_2^2x_3}
\le 0,
\qquad&
g_4(x) &=
\frac{x_1+x_2}{1.5}-1
\le 0.
\end{alignedat}
\end{equation}
Here, $g_1$ imposes the required deflection condition; $g_2$ limits the allowable shear stress; $g_3$ imposes the natural frequency requirement; and $g_4$ restricts the overall spring diameter.

Constraint violations in the numerical experiments are handled using a quadratic exterior penalty:
\begin{equation}
F(x)=f(x)
+\rho\sum_{j=1}^{4}
\left[\max\{0,g_j(x)\}\right]^2,
\label{eq:classical_penalty}
\end{equation}
with $\rho=10^6$. Thus, designs that are not feasible incur an additional cost proportional to the squared magnitude of their constraint violations.

\subsection{$MPR$ parameterised robust TCSD formulation}
\label{subsec:MPR}

The proposed $MPR$ parameterisation extends the classical TCSD defined in
the preceding subsection. It retains the original 3-dimensional
decision space and the nominal objective function given in
Eq.~\eqref{eq:classical_spring_objective}. Robustness is introduced solely through controlled modifications of the feasibility conditions.This creates a family of related TCSD instances without increasing the problem
dimension.

Three binary configuration parameters are introduced: $C=[M,P,R]\in\{0,1\}^3$.
These parameters are fixed prior to optimisation and are not decision variables.
Their eight combinations define the $MPR$ configurations listed
in Table~\ref{tab:MPR_configs}.
\begin{table}[ht]
\centering
\caption{Configurations of the proposed $MPR$-parameterized TCSD benchmark.}
\label{tab:MPR_configs}
\renewcommand{\arraystretch}{1.08}
\setlength{\tabcolsep}{3.8pt}
\begin{tabular}{@{}c l c c c l@{}}
\toprule
No. & Configuration & $M$ & $P$ & $R$ & Comparison purpose \\
\midrule
1. & $C(000)$ & 0 & 0 & 0 & Nominal TCSD; all $MPR$ mechanisms disabled\\
2. & $C(100)$ & 1 & 0 & 0 & Material variability only \\
3. & $C(010)$ & 0 & 1 & 0 & Geometric perturbation only \\
4. & $C(001)$ & 0 & 0 & 1 & Requirements margin only \\
5. & $C(110)$ & 1 & 1 & 0 & Material and geometric effects  \\
6. & $C(101)$ & 1 & 0 & 1 & Material and requirements effects  \\
7. & $C(011)$ & 0 & 1 & 1 & Geometric and requirements effects \\
8. & $C(111)$ & 1 & 1 & 1 & All $MPR$ mechanisms active \\
\bottomrule
\end{tabular}
\end{table}
The eight configurations form a complete two-level $2^3$ factorial
arrangement. As noted by Telford in \cite{JKT2007}, such a design enables
the individual and combined effects of $M$, $P$, and $R$ to be examined
under directly comparable conditions. Configuration $C(000)$ represents the nominal TCSD,
whereas $C(111)$ activates all three $MPR$ mechanisms.

The material and functional scales are supported by Cadet and Paredes in \cite{GC2024}, whereas the geometric perturbation is justified separately below. These suggestions translate into our formulation as: 
\[\tau_M=0.07,\qquad \tau_P=0.02,\qquad \eta_R=0.10.\] 
The authors indicate approximately $\pm7\%$ for mechanical material parameters,
approximately $\pm2\%$ for dimensional parameters, and approximately
$\pm10\%$ for functional parameters. 
In their subsequent work, Cadet and Paredes in \cite{GC2026} further confirm
the importance of material variability, manufacturing uncertainty, and
specification tolerances in robust spring design.
Their mathematical incorporation into the classical TCSD constitutes the
proposed $MPR$ parameterisation.

Material variability is introduced through $\alpha_M=\tau_M M$.
For $M=0$, the nominal form of the constraints is recovered. When $M=1$,
a relative material variability amplitude of $7\%$ is activated.
No additional material variable is introduced. Instead, the effect of
material variability is incorporated directly into the classical
constraints:
\begin{equation}
\label{eq:M_constraints}
\begin{alignedat}[c]{2}
g_1(x; M)
&=
1-
\frac{1}{1+\alpha_M}
\frac{x_2^3x_3}{71785x_1^4},
\quad&
g_2(x; M)
&=
\frac{1}{1-\alpha_M}
\left[
\frac{4x_2^2-x_1x_2}
     {12566\left(x_2x_1^3-x_1^4\right)}
+
\frac{1}{5108x_1^2}
\right]-1,
\\[0.8ex]
g_3(x; M)
&=
1-
\sqrt{1-\alpha_M}
\frac{140.45x_1}{x_2^2x_3},
\quad&
g_4(x; M)
&=
g_4(x).
\end{alignedat}
\end{equation}

The modifications in Eq.~\eqref{eq:M_constraints} follow the physical
dependencies represented by the classical TCSD constraints in
Eq.~\eqref{eq:classical_constraints}. Spring
deflection is inversely proportional to the shear modulus $G$.
An increase in $G$ therefore reduces spring compliance and is contrary
to the minimum-deflection requirement represented by $g_1$.
Accordingly, the adverse material state is represented by $G=G_0(1+\alpha_M)$,
where $G_0$ denotes the nominal shear modulus. Substituting this into the
deflection relation introduces the factor $1/(1+\alpha_M)$ in $g_1(x; M)$. 
The numerical coefficient $71785$ belongs
to the nominal benchmark formulation and incorporates  the nominal constants
associated with this deflection relation. The MPR modification therefore
acts on the material-dependent part of the constraint while preserving
the classical benchmark structure.

In $g_2(x; M)$, the admissible stress level embedded in the classical
constraint is reduced by the factor $1-\alpha_M$. This produces the
multiplier $1/(1-\alpha_M)$, with $\alpha_M$ acting as a downward
perturbation of the material tensile strength $R_m$ (i.e., a reduction
in the admissible strength level). This treatment is an $MPR$ modelling
assumption.

The variability of mechanical material properties reported by Cadet
and Paredes in \cite{GC2024} motivates the adopted amplitude.
It neither reproduces their material model nor introduces $R_m$ as an
explicit material variable.

For the natural frequency $f_r$ (the free-vibration frequency), and for
fixed geometry and density, $f_r\propto\sqrt{G}$ (i.e., $f_r$ scales
with $\sqrt{G}$).  Therefore, a decrease in $G$ reduces $f_r$. Accordingly,
$g_3(x; M)$ assumes $G=G_0(1-\alpha_M)$, where $G_0$ is the nominal shear modulus. 
This substitution produces the
factor $\sqrt{1-\alpha_M}$ in $g_3(x; M)$. Constraint $g_4(x; M)$ is purely
geometric and therefore remains independent of $M$.

The adverse direction of material variability is selected separately for
each mechanically dependent constraint. For $g_1(x; M)$, it corresponds to
a higher value of $G$, whereas for $g_3(x; M)$ it corresponds to a lower
value. Consequently, $M$ does not represent a single coherent material
realisation.  Instead, it defines an adverse envelope of the limiting 
material states associated with the individual requirements, as defined 
by the constraints.
This differs from the extreme spring approach described by Cadet and
Paredes in \cite{GC2024}, in which the parameter sets define coherent
softest and stiffest spring realisations.

The parameter $P$ represents geometric variability related to manufacturing.
For a nominal design vector $x$, the perturbed geometry is defined as follows:
$$x^{(\boldsymbol{s})}
=
x\odot
\left(
\mathbf{1}+\tau_P P\boldsymbol{s}
\right),
\qquad
\boldsymbol{s}\in\{-1,+1\}^3,$$
where $\odot$ denotes element-wise multiplication. For $P=1$ and when 
the perturbation parameter is set to $0.02$, each coordinate takes the value 
of either $0.98x_i$ or $1.02x_i$, where $i$ ranges from 1 to 3. The corresponding 
finite perturbation set is therefore:
$$\mathcal{U}_{P}(x)
=
\left\{
x\odot
\left(
\mathbf{1}+0.02\boldsymbol{s}
\right)
:
\boldsymbol{s}\in\{-1,+1\}^3
\right\}.$$
It contains all eight corner geometries associated with the same nominal
design. These perturbed geometries are only used for constraint evaluation
and do not constitute additional decision variables.

The construction is related 
to the extreme spring treatment described by
Cadet and Paredes in \cite{GC2024} for representing parameter variability.
Using the monotonic relationships, they construct coherent softest and
stiffest spring realisations, denoted by: $\{x\}^{L}$ and $\{x\}^{U}$,
respectively. However, $MPR$ does not apply this monotonicity-based reduction. 
Instead, all eight corner geometries in $\mathcal{U}_{P}(x)$, are evaluated explicitly without 
reducing it to two coherent extreme springs.

The amplitude $\tau_P=0.02$ is applied to all three TCSD decision variables. The dimensional tolerance scale reported by Cadet and Paredes in \cite{GC2024} is used for $x_1$ and $x_2$. For $x_3$, the same level is introduced as a modelling choice, since their industrial cases assume an active coil manufacturing variability of $0.1\%$. The broader coil-count variation of up to $\pm5\%$ reported by Oberg et al. in \cite{EO2004} provides an upper practical reference.
Thus, the selected $2\%$ level is moderate and ensures a uniform relative perturbation scale across the continuous TCSD decision space.

Rather than representing an additional uncertain physical quantity, the parameter $R$ defines a uniform margin in the normalised constraint space:
\begin{equation}
\widetilde g_j
\left(
x^{(\boldsymbol{s})},M,R
\right)
=
g_j
\left(
x^{(\boldsymbol{s})}; M
\right)
+\eta_R R,
\qquad j=1,\ldots,4.
\label{eq:R_margin}
\end{equation}
According to Eq.~(\ref{eq:R_margin}), $R=0$ leaves the original requirements unchanged.
For $R=1$ and $\eta_R=0.10$, feasibility requires:
$g_j
\left(
x^{(\boldsymbol{s})}; M
\right)
\leq -0.10$.
Thus, a normalised margin of $0.10$ is enforced within the feasible region.

The functional-tolerance scale reported by Cadet and Paredes in \cite{GC2024}
motivates the choice of parameter, $\eta_R=0.10$. They associate approximately
$\pm10\%$ with functional parameters. Their industrial examples apply this scale to operating loads and stiffness. However, the $MPR$ mechanism uses this scale differently. In the $MPR$ formulation, this scale represents a deterministic margin of requirements. It does not directly affect a functional quantity or represent a probabilistic reliability measure.

For a fixed configuration $C$, the resulting robust problem can be expressed as follows:
\begin{equation}
\begin{aligned}
\min_{x\in\Omega}\quad & f(x),\\
\text{subject to}\quad &
\widetilde g_j
\left(
x^{(\boldsymbol{s})},M,R
\right)
\leq 0,
\\[-0.2ex]
& j=1,\ldots,4,\qquad
\boldsymbol{s}\in\{-1,+1\}^{3}.
\end{aligned}
\label{eq:MPR_problem}
\end{equation}
Here, $\Omega$ denotes the decision domain defined in
Eq.~\eqref{eq:DzOmega}. For $P=0$, all perturbed geometries reduce to the nominal geometry.
For $P=1$, the constraints must be satisfied by all eight corner geometries
in $\mathcal{U}_{P}(x)$. This is consistent
with the deterministic robust-optimisation framework of Li, et al. in \cite{CL2024}, Gorissen, et al. in \cite{BLG2015}, Bertsimas, et al. in \cite{DB2011}, Ben-Tal, et al. in \cite{AB-T2010},
because all admissible realisations are evaluated explicitly and no
probability distribution is required.

The objective function, $f(x)$, is evaluated at the nominal geometry. The $MPR$ only modifies the feasibility conditions, leaving the classical mass objective unchanged. This preserves direct comparability with the nominal TCSD.
Therefore, any increase in the optimal value of $f$ after activating $MPR$ can be interpreted as the cost of increased robustness.

The degree of infeasibility is quantified as follows:
\begin{equation}
V(x;M,P,R)
=
\max_{\boldsymbol{s}\in\{-1,+1\}^{3}}
\sum_{j=1}^{4}
\left[
\max\left(
0,
\widetilde g_j
\left(
x^{(\boldsymbol{s})},M,R
\right)
\right)
\right]^2.
\label{eq:Viol}
\end{equation}
The positive-part operator excludes satisfied constraints. 
The sum aggregates the squared constraint violations for each perturbed
geometry. The outer maximum then selects the worst-case geometry.
For $P=0$, the maximum is trivial because all perturbed geometries reduce
to the nominal geometry. For $P=1$, the maximum is the corner geometry with
the largest aggregate violation. Consequently, $V(x;M,P,R)=0$ if and only if all four constraints are
satisfied for every geometry in the finite perturbation set.
For numerical optimisation, Eq.~\eqref{eq:MPR_problem} is represented by
the penalised objective:
\begin{equation}
F(x;M,P,R)=f(x)+10^{6}V(x;M,P,R).
\label{eq:MPR_penalty}
\end{equation}
The external penalty that Cadet and Paredes in \cite{GC2024} define in their
Eq.~(5) similarly uses squared positive normalised constraint violations.
Their subsequent objective functions apply this penalty to the two
extreme springs. In $MPR$, the TCSD constraints are already
dimensionless. Equation~\eqref{eq:Viol} additionally takes the maximum
aggregate violation over the finite perturbation set
$\mathcal{U}_{P}(x)$. The aggregation rule is therefore different.
The coefficient $10^{6}$ is retained from the quadratic exterior penalty
used for the nominal TCSD in Eq.~\eqref{eq:classical_penalty}. This is
independent of the value $\rho=1000$ used by Cadet and Paredes in \cite{GC2024}.

\section{Experimental protocol}

The experiments were carried out in MATLAB R2023b using a unified
computational framework. For each baseline optimiser, the implementation
was taken from source code released by its authors or from the resources
cited in the original publication. For GA (see, e.g., Deb and Kumar in \cite{KD1995} and Deb in \cite{KD2000}), the \texttt{Real\_Coded\_GA} MATLAB implementation by Zhan in
\cite{DZ2026} was used. In each case (except CMA-ES, e.g. Hansen et al. in  \cite{NH2004}), 
the corresponding hybrid variant was created by adding the
predicted candidate operator described in
Section~\ref{sec:predicted_candidate_operator}. The numerical results
presented here originate from the experiments performed for this study.
The study included 15 optimisation methods evaluated across eight $MPR$
configurations.

Each algorithm-configuration pair was tested in 30 independent runs.
The problem dimension was $dim=3$. The evaluation budget was
$FE_{\max}=30000$ function evaluations per run. Methods with a fixed population size used 20 individuals. 
This choice is consistent with the findings of Chen et al. in \cite{TCh2012}, who found that larger populations 
did not necessarily lead to better performance under limited evaluation budgets.
L-SHADE and hL-SHADE retained linear population size reduction, with an initial
population of $18\times dim=54$. 

One function evaluation corresponds to
a complete assessment of one candidate under the current $MPR$
configuration. Algorithm-specific parameter settings are listed in
\textbf{Table~\ref{tab:Parameters_Algo}.}

\clearpage
\begin{table}[ht]
\centering
\caption{Hyperparameter settings of the investigated base algorithms.}
\label{tab:Parameters_Algo}
\renewcommand{\arraystretch}{1.2} 
\resizebox{!}{2.4cm}{%
\setlength{\tabcolsep}{6pt} 
\begin{tabular}{@{} l l l @{}}
\toprule
No. &  Algorithm & Hyperparameter setting \\
\midrule
1. & CMA-ES \cite{NH2004}                        &  $\lambda=20$, $\mu=10$, $\sigma^{(0)}=0.3(U-L)$ ($L/U$:                          lower/upper bounds). \\ 
2. & DE \cite{RS1997} /hDE              & $DE/rand/1/bin:$ $NP=20$, $F=0.5$,$CR=0.9$.\\
3. & L-SHADE \cite{RT2014}/hL-SHADE     & $N_{\mathrm{init}}=18\cdot dim$, $N_{\min}=4$, $r_{\mathrm{arc}}=2.6$, $p=0.11$, $H=6$.\\
4. & MPA \cite{AF2020}/hMPA             & $P=0.5$, $\mathrm{FADs}=0.2$, $\alpha=1.5$; 3-equal search phases.\\
5. & SPSO2011 \cite{MC2012}/hSPSO2011    & $w=1/(2\ln 2)$, $c_1=c_2=0.5+\ln 2$, $K=3$.\\
6. & WOA \cite{SM2016}/hWOA             & $a:2\rightarrow 0$; random search updates. \\
7. & EO \cite{AFeo2020}/hEO             & $a_1=2$, $a_2=1$, $GP=0.5$, $V=1$. \\
8. & GA \cite{KD1995},\cite{KD2000}/hGA & $k=2$, $p_c=1$, $\eta_c=1$,$p_m=1/D$, $\eta_m=1$. \\    
\bottomrule
\end{tabular}
}
\end{table}

All computations, analyses and figure preparation were performed on a Dell 
computer running Windows 10 Home 22H2 with an Intel Core i7-7500U processor 
at 2.70 GHz and 8 GB of RAM.

\section{Results}

\subsection{3D objective-constraint violation landscapes}

The objective and the violation of the constraints are analysed separately. This follows the constrained-landscape perspective of Vodopija et al. in \cite{AV2022}, where objective information and the overall extent of the constraint violation are treated as distinct landscape components. Here, constraint violation is quantified by $V(x; M, P, R)$ in 
Eq. (\ref{eq:Viol}). The penalised objective is not used.
A common Latin Hypercube Sample (LHS) of $100000$ points spanning the decision domain in Eq.~(\ref{eq:DzOmega}) is evaluated for all eight $MPR$ configurations listed in Table~\ref{tab:MPR_configs} 
(e.g., Santner et al. in \cite{TJS2018}, Mayda in \cite{MM2018} and Komeilizadeh et al. in \cite{KK2023} for discussions of LHS-based space-filling designs). Positive $V$ values are pooled across configurations. Two common low violation levels are obtained from the first and fifth percentiles of the pooled $\log_{10}\left(V\right)$ values. These levels remain fixed for all configurations.

First, a common visualisation window is determined from the union of low-violation regions across all configurations. This window is then used for the preliminary evaluation. The violation field is then evaluated directly on the same $171\times171\times171$ grid (chosen as a compromise between geometric resolution and computational cost) for each configuration. The boundary of the feasible region is reconstructed from the feasibility field. The colour of this boundary represents the nominal objective $f$ in Eq.~(\ref{eq:classical_spring_objective}). A common objective colour scale is used in all panels.

\begin{figure}[ht]
\noindent\hspace*{-0.166\textwidth}%
\begin{subfigure}{1.35\textwidth}
    \centering
    \includegraphics[width=\linewidth, keepaspectratio]{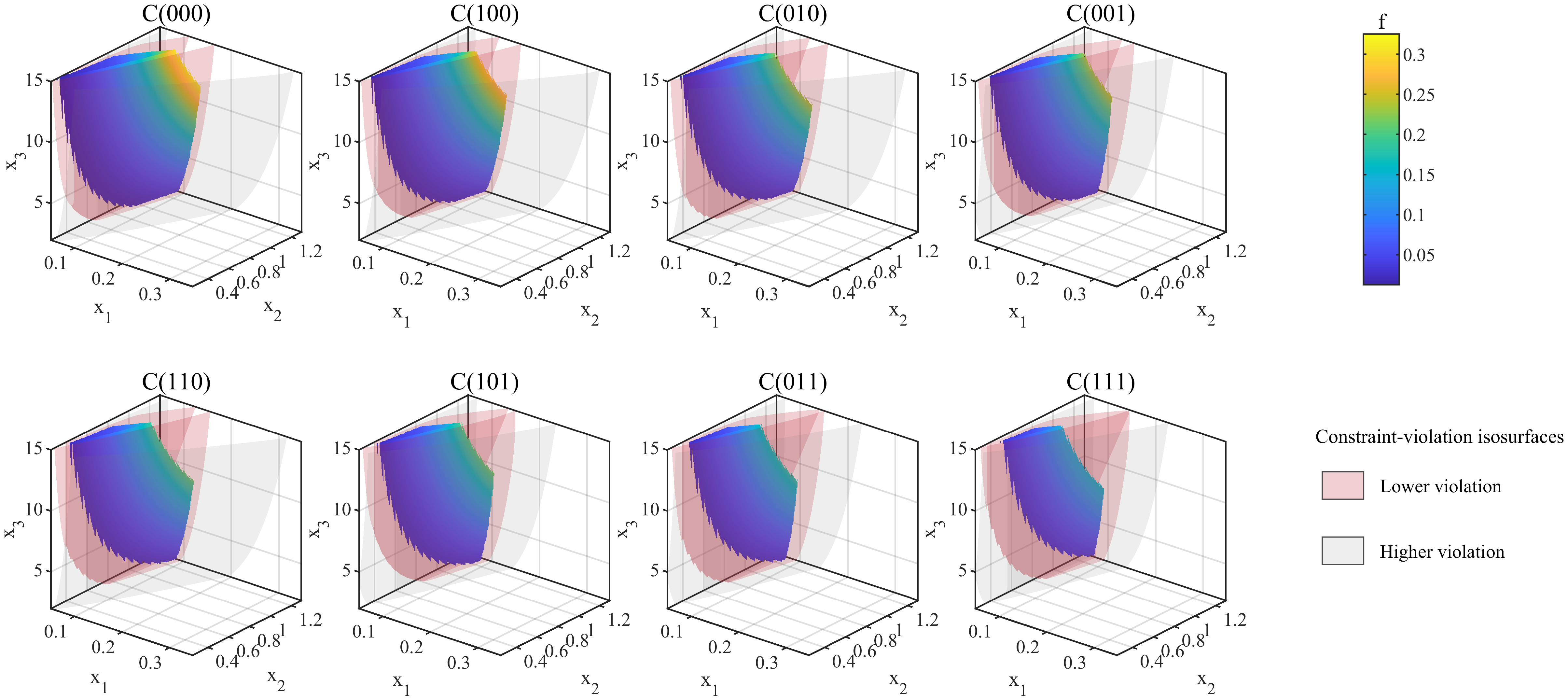}
\end{subfigure}
\caption{Objective-violation geometry across the MPR configurations.}
\label{fig1}
\end{figure}

Two semi-transparent isosurfaces represent lower and higher constraint violations. This adapts the 3-dimensional isosurface principle of Sch\"apermeier et al. in \cite{LS2021} to the $MPR$ violation field. The resulting landscapes are shown in Fig.~\ref{fig1}.
\clearpage
The coloured surface represents the boundary of the feasible region, with colour indicating the nominal objective value $f(x)$. The reddish and grey semi-transparent isosurfaces denote lower and higher constraint-violation levels, $V=0.146427$ and $V=0.933632$, respectively. The same violation levels and objective colour scale are used for all configurations.

Figure~\ref{fig1} shows that the $MPR$ mechanisms substantially alter the geometry and location of the feasible region. Combined configurations produce stronger changes than the nominal case, $C(000)$, with $C(111)$ exhibiting the most restrictive structure. The surrounding violation isosurfaces change accordingly. These results demonstrate that $MPR$ primarily reshapes feasibility, while the nominal objective function remains unchanged.

\subsection{Table}

The eight $MPR$ configurations were evaluated using all 15 algorithms. Table~\ref{tab:MPR_all_results} reports the resulting design variables and performance statistics.

\clearpage
\begin{table*}[p]
\centering
\caption{Complete optimization results for the $8$ configurations of the 3-variable $MPR$-parameterized TCSD benchmark.}
\label{tab:MPR_all_results}
\begingroup

\scriptsize
\setlength{\tabcolsep}{1.5pt}
\renewcommand{\arraystretch}{1.02}
\centerline{%
\resizebox{1.2\textwidth}{!}{%
\begin{tabular}{@{}c D{.}{.}{-1} D{.}{.}{-1} D{.}{.}{-1} D{.}{.}{-1} D{.}{.}{-1} D{.}{.}{-1} D{.}{.}{-1} @{\hspace{32pt}} l D{.}{.}{-1}@{}}
\toprule

\multicolumn{1}{c}{{\small Configuration}} &
\multicolumn{1}{c}{{\small $x_1$}} &
\multicolumn{1}{c}{{\small $x_2$}} &
\multicolumn{1}{c}{{\small $x_3$}} &
\multicolumn{1}{c}{{\small $F_{best}$}} &
\multicolumn{1}{c}{{\small $\overline{F}$}} &
\multicolumn{1}{c}{{\small $\mathrm{SD}(F)$}} &
\multicolumn{1}{c}{{\small $\overline{f}$}} &
\multicolumn{1}{c}{{\small $\overline{V}$}} &
\multicolumn{1}{c}{{\small FR}} \\

\midrule
\multicolumn{10}{@{}l}{\textbf{CMA-ES} \cite{NH2004}}\\[-0.8pt]
\addlinespace[0.2pt]
$C(000)$ & 0.051685 & 0.356631 & 11.294035 & 0.012665 & 0.012665 & 0.000000 & 0.012665 & 2.029E-16 & 1.000 \\*
$C(100)$ & 0.054201 & 0.384559 & 11.656624 & 0.015429 & 0.015429 & 0.000000 & 0.015429 & 3.317E-16 & 1.000 \\*
$C(010)$ & 0.054500 & 0.391712 & 12.365485 & 0.016714 & 5486.140223 & 1859.970064 & 0.013860 & 5.486E-3 & 0.100 \\*
$C(001)$ & 0.055361 & 0.397447 & 11.814601 & 0.016828 & 0.016828 & 0.000000 & 0.016828 & 3.959E-16 & 1.000 \\*
$C(110)$ & 0.057155 & 0.422235 & 12.778568 & 0.020385 & 16539.613672 & 11012.807848 & 0.015539 & 1.654E-2 & 0.300 \\*
$C(101)$ & 0.058063 & 0.428632 & 12.194374 & 0.020512 & 0.020512 & 0.000000 & 0.020512 & 5.985E-16 & 1.000 \\*
$C(011)$ & 0.058384 & 0.436422 & 12.952926 & 0.022244 & 28803.680680 & 9765.354778 & 0.014425 & 2.880E-2 & 0.100 \\*
$C(111)$ & 0.061261 & 0.471184 & 13.350036 & 0.027144 & 62888.179718 & 17094.895388 & 0.014455 & 6.289E-2 & 0.067 \\
\addlinespace[1.0pt]
\midrule
\addlinespace[0.4pt]
\multicolumn{10}{@{}l}{\textbf{DE} \cite{RS1997}}\\[-0.8pt]
\addlinespace[0.2pt]
$C(000)$ & 0.051689 & 0.3567180 & 11.288966 & 0.012665 & 0.012694 & 0.000079 & 0.012694 & 1.664E-16 & 1.000 \\*
$C(100)$ & 0.054203 & 0.384596 & 11.654556 & 0.015429 & 0.015762 & 0.001208 & 0.015762 & 2.526E-16 & 1.000 \\*
$C(010)$ & 0.054494 & 0.391559 & 12.374363 & 0.016714 & 0.016984 & 0.001195 & 0.016984 & 3.913E-16 & 1.000 \\*
$C(001)$ & 0.055362 & 0.397460 & 11.813891 & 0.016828 & 0.016844 & 0.000037 & 0.016844 & 3.652E-16 & 1.000 \\*
$C(110)$ & 0.057158 & 0.422301 & 12.774881 & 0.020385 & 0.020439 & 0.000140 & 0.020439 & 4.569E-16 & 1.000 \\*
$C(101)$ & 0.058065 & 0.428673 & 12.192246 & 0.020512 & 0.020525 & 0.000027 & 0.020525 & 5.449E-16 & 1.000 \\*
$C(011)$ & 0.058391 & 0.436600 & 12.943268 & 0.022244 & 2133.626674 & 8119.691493 & 0.021766 & 2.134E-3 & 0.933 \\*
$C(111)$ & 0.061256 & 0.471047 & 13.357136 & 0.027144 & 4492.038202 & 17094.895345 & 0.026398 & 4.492E-3 & 0.933 \\
\addlinespace[1.0pt]
\midrule
\addlinespace[0.4pt]
\multicolumn{10}{@{}l}{\textbf{hDE}}\\[-0.8pt]
\addlinespace[0.2pt]
$C(000)$ & 0.051688 & 0.356696 & 11.290262 & 0.012665 & 0.012742 & 0.000199 & 0.012742 & 1.576E-16 & 1.000 \\*
$C(100)$ & 0.054204 & 0.384625 & 11.652948 & 0.015429 & 0.015647 & 0.001096 & 0.015647 & 2.390E-16 & 1.000 \\*
$C(010)$ & 0.054494 & 0.391562 & 12.374185 & 0.016714 & 0.016733 & 0.000048 & 0.016733 & 6.701E-16 & 1.000 \\*
$C(001)$ & 0.055360 & 0.397418 & 11.816163 & 0.016828 & 0.016858 & 0.000053 & 0.016858 & 3.291E-16 & 1.000 \\*
$C(110)$ & 0.057160 & 0.422363 & 12.771462 & 0.020385 & 0.020950 & 0.001951 & 0.020950 & 4.525E-16 & 1.000 \\*
$C(101)$ & 0.058067 & 0.428722 & 12.189671 & 0.020512 & 0.020564 & 0.000096 & 0.020564 & 4.820E-16 & 1.000 \\*
$C(011)$ & 0.058407 & 0.437009 & 12.921134 & 0.022244 & 0.022532 & 0.000826 & 0.022532 & 6.915E-16 & 1.000 \\*
$C(111)$ & 0.061256 & 0.471049 & 13.357029 & 0.027144 & 0.027513 & 0.001259 & 0.027513 & 1.034E-15 & 1.000 \\
\addlinespace[1.0pt]
\midrule
\addlinespace[0.4pt]
\multicolumn{10}{@{}l}{\textbf{L-SHADE} \cite{RT2014}}\\[-0.8pt]
\addlinespace[0.2pt]
$C(000)$ & 0.051689 & 0.356718 & 11.288966 & 0.012665 & 0.012665 & 0.000000 & 0.012665 & 5.294E-16 & 1.000 \\*
$C(100)$ & 0.054203 & 0.384596 & 11.654556 & 0.015429 & 0.015429 & 0.000000 & 0.015429 & 2.886E-16 & 1.000 \\*
$C(010)$ & 0.054494 & 0.391562 & 12.374243 & 0.016714 & 0.016714 & 0.000000 & 0.016714 & 5.506E-16 & 1.000 \\*
$C(001)$ & 0.055363 & 0.397482 & 11.812665 & 0.016828 & 0.016828 & 0.000000 & 0.016828 & 3.657E-16 & 1.000 \\*
$C(110)$ & 0.057160 & 0.422363 & 12.771463 & 0.020385 & 0.020385 & 0.000002 & 0.020385 & 8.809E-16 & 1.000 \\*
$C(101)$ & 0.058064 & 0.428649 & 12.193499 & 0.020512 & 0.020512 & 0.000001 & 0.020512 & 8.614E-16 & 1.000 \\*
$C(011)$ & 0.058391 & 0.436600 & 12.943269 & 0.022244 & 0.022245 & 0.000003 & 0.022245 & 1.983E-15 & 1.000 \\*
$C(111)$ & 0.061256 & 0.471047 & 13.357134 & 0.027144 & 0.027144 & 0.000001 & 0.027144 & 2.828E-15 & 1.000 \\
\addlinespace[1.0pt]
\midrule
\addlinespace[0.4pt]
\multicolumn{10}{@{}l}{\textbf{hL-SHADE}}\\[-0.8pt]
\addlinespace[0.2pt]
$C(000)$ & 0.051689 & 0.356718 & 11.288966 & 0.012665 & 0.012665 & 0.000001 & 0.012665 & 3.634E-16 & 1.000 \\*
$C(100)$ & 0.054203 & 0.384596 & 11.654556 & 0.015429 & 0.015429 & 0.000000 & 0.015429 & 5.055E-16 & 1.000 \\*
$C(010)$ & 0.054494 & 0.391562 & 12.374243 & 0.016714 & 0.016714 & 0.000001 & 0.016714 & 2.278E-15 & 1.000 \\*
$C(001)$ & 0.055363 & 0.397482 & 11.812666 & 0.016828 & 0.016828 & 0.000001 & 0.016828 & 5.145E-16 & 1.000 \\*
$C(110)$ & 0.057160 & 0.422363 & 12.771463 & 0.020385 & 0.020385 & 0.000002 & 0.020385 & 1.351E-15 & 1.000 \\*
$C(101)$ & 0.058064 & 0.428649 & 12.193499 & 0.020512 & 0.020512 & 0.000002 & 0.020512 & 3.091E-15 & 1.000 \\*
$C(011)$ & 0.058391 & 0.436600 & 12.943269 & 0.022244 & 0.022245 & 0.000002 & 0.022245 & 4.625E-15 & 1.000 \\*
$C(111)$ & 0.061256 & 0.471047 & 13.357134 & 0.027144 & 0.027145 & 0.000003 & 0.027145 & 6.627E-15 & 1.000 \\
\addlinespace[1.0pt]
\midrule
\addlinespace[0.4pt]
\multicolumn{10}{@{}l}{\textbf{MPA} \cite{AF2020}}\\[-0.8pt]
\addlinespace[0.2pt]
$C(000)$ & 0.051688 & 0.356692 & 11.290466 & 0.012665 & 0.012665 & 0.000000 & 0.012665 & 2.026E-16 & 1.000 \\*
$C(100)$ & 0.054201 & 0.384536 & 11.657925 & 0.015429 & 0.015429 & 0.000000 & 0.015429 & 3.266E-16 & 1.000 \\*
$C(010)$ & 0.054490 & 0.391475 & 12.379292 & 0.016714 & 0.016714 & 0.000000 & 0.016714 & 5.533E-16 & 1.000 \\*
$C(001)$ & 0.055358 & 0.397349 & 11.819964 & 0.016828 & 0.016828 & 0.000000 & 0.016828 & 3.821E-16 & 1.000 \\*
$C(110)$ & 0.057165 & 0.422479 & 12.765015 & 0.020385 & 0.020385 & 0.000000 & 0.020385 & 7.498E-16 & 1.000 \\*
$C(101)$ & 0.058062 & 0.428592 & 12.196460 & 0.020512 & 0.020512 & 0.000000 & 0.020512 & 8.286E-16 & 1.000 \\*
$C(011)$ & 0.058389 & 0.436561 & 12.945373 & 0.022244 & 0.022244 & 0.000000 & 0.022244 & 7.967E-16 & 1.000 \\*
$C(111)$ & 0.061260 & 0.471136 & 13.352491 & 0.027144 & 0.027144 & 0.000000 & 0.027144 & 1.244E-15 & 1.000 \\
\addlinespace[1.0pt]
\midrule
\addlinespace[0.4pt]
\multicolumn{10}{@{}l}{\textbf{hMPA}}\\[-0.8pt]
\addlinespace[0.2pt]
$C(000)$ & 0.051694 & 0.356844 & 11.281565 & 0.012665 & 0.012665 & 0.000000 & 0.012665 & 1.719E-16 & 1.000 \\*
$C(100)$ & 0.054203 & 0.384592 & 11.654801 & 0.015429 & 0.015429 & 0.000000 & 0.015429 & 2.523E-16 & 1.000 \\*
$C(010)$ & 0.054496 & 0.391609 & 12.371469 & 0.016714 & 0.016714 & 0.000000 & 0.016714 & 7.592E-16 & 1.000 \\*
$C(001)$ & 0.055366 & 0.397559 & 11.808443 & 0.016828 & 0.016828 & 0.000000 & 0.016828 & 8.343E-16 & 1.000 \\*
$C(110)$ & 0.057165 & 0.422472 & 12.765388 & 0.020385 & 0.020385 & 0.000000 & 0.020385 & 4.288E-16 & 1.000 \\*
$C(101)$ & 0.058071 & 0.428842 & 12.183380 & 0.020512 & 0.020512 & 0.000000 & 0.020512 & 4.475E-16 & 1.000 \\*
$C(011)$ & 0.058387 & 0.436504 & 12.948468 & 0.022244 & 0.022244 & 0.000000 & 0.022244 & 9.410E-16 & 1.000 \\*
$C(111)$ & 0.061258 & 0.471088 & 13.354997 & 0.027144 & 0.027144 & 0.000001 & 0.027144 & 1.246E-15 & 1.000 \\
\addlinespace[1.0pt]
\midrule
\addlinespace[0.4pt]
\multicolumn{10}{@{}l}{\textbf{SPSO2011} \cite{MC2012}}\\[-0.8pt]
\addlinespace[0.2pt]
$C(000)$ & 0.052082 & 0.366256 & 10.752151 & 0.012669 & 0.012820 & 0.000289 & 0.012820 & 9.144E-18 & 1.000 \\*
$C(100)$ & 0.054388 & 0.389161 & 11.403985 & 0.015430 & 0.015591 & 0.000200 & 0.015591 & 1.781E-17 & 1.000 \\*
$C(010)$ & 0.054413 & 0.389575 & 12.490413 & 0.016714 & 0.016977 & 0.000292 & 0.016977 & 2.770E-15 & 1.000 \\*
$C(001)$ & 0.055844 & 0.409537 & 11.183448 & 0.016837 & 0.017173 & 0.000490 & 0.017173 & 7.899E-17 & 1.000 \\*
$C(110)$ & 0.057706 & 0.436293 & 12.047392 & 0.020408 & 0.020784 & 0.000302 & 0.020784 & 3.352E-16 & 1.000 \\*
$C(101)$ & 0.058001 & 0.427048 & 12.280774 & 0.020516 & 0.020805 & 0.000296 & 0.020805 & 7.687E-16 & 1.000 \\*
$C(011)$ & 0.058183 & 0.431285 & 13.238329 & 0.022248 & 0.022465 & 0.000224 & 0.022465 & 4.748E-17 & 1.000 \\*
$C(111)$ & 0.061250 & 0.470888 & 13.366268 & 0.027146 & 0.027572 & 0.000393 & 0.027572 & 7.141E-17 & 1.000 \\
\bottomrule
\end{tabular}
}
}
\endgroup
\end{table*}

\begin{table*}[p]
\centering
\begingroup
\scriptsize
\setlength{\tabcolsep}{1.5pt}
\renewcommand{\arraystretch}{1.02}
\noindent\textit{Table~\ref{tab:MPR_all_results} (continued).}\par\vspace{2pt}
\centerline{%
\resizebox{1.2\textwidth}{!}{%
\begin{tabular}{@{}c D{.}{.}{-1} D{.}{.}{-1} D{.}{.}{-1} D{.}{.}{-1} D{.}{.}{-1} D{.}{.}{-1} D{.}{.}{-1} @{\hspace{32pt}} l D{.}{.}{-1}@{}}

\toprule
\multicolumn{1}{c}{{\small Configuration}} &
\multicolumn{1}{c}{{\small $x_1$}} &
\multicolumn{1}{c}{{\small $x_2$}} &
\multicolumn{1}{c}{{\small $x_3$}} &
\multicolumn{1}{c}{{\small $F_{best}$}} &
\multicolumn{1}{c}{{\small $\overline{F}$}} &
\multicolumn{1}{c}{{\small $\mathrm{SD}(F)$}} &
\multicolumn{1}{c}{{\small $\overline{f}$}} &
\multicolumn{1}{c}{{\small $\overline{V}$}} &
\multicolumn{1}{c}{{\small FR}} \\
\midrule
\multicolumn{10}{@{}l}{\textbf{hSPSO2011}}\\[-0.8pt]
\addlinespace[0.2pt]
$C(000)$ & 0.051803 & 0.359460 & 11.130029 & 0.012666 & 0.012841 & 0.000255 & 0.012841 & 1.262E-15 & 1.000 \\*
$C(100)$ & 0.053564 & 0.369061 & 12.578307 & 0.015437 & 0.015658 & 0.000274 & 0.015658 & 3.342E-17 & 1.000 \\*
$C(010)$ & 0.054775 & 0.398560 & 11.978234 & 0.016715 & 0.017089 & 0.000380 & 0.017089 & 3.111E-16 & 1.000 \\*
$C(001)$ & 0.054785 & 0.383269 & 12.635726 & 0.016836 & 0.017144 & 0.000437 & 0.017144 & 4.827E-17 & 1.000 \\*
$C(110)$ & 0.057279 & 0.425365 & 12.606989 & 0.020385 & 0.020699 & 0.000276 & 0.020699 & 2.357E-17 & 1.000 \\*
$C(101)$ & 0.058140 & 0.430582 & 12.094199 & 0.020514 & 0.020897 & 0.000461 & 0.020897 & 1.888E-16 & 1.000 \\*
$C(011)$ & 0.058069 & 0.428421 & 13.399453 & 0.022246 & 0.022599 & 0.000332 & 0.022599 & 4.683E-17 & 1.000 \\*
$C(111)$ & 0.061757 & 0.484260 & 12.711379 & 0.027171 & 0.027653 & 0.000677 & 0.027653 & 3.129E-14 & 1.000 \\
\addlinespace[1.0pt]
\midrule
\addlinespace[0.4pt]
\multicolumn{10}{@{}l}{\textbf{WOA} \cite{SM2016}}\\[-0.8pt]
\addlinespace[0.2pt]
C(000) & 0.053332 & 0.397552 & 9.242998 & 0.012713 & 0.013746 & 0.001136 & 0.013746 & 3.162E-16 & 1.000 \\
C(100) & 0.053358 & 0.364124 & 12.896290 & 0.015443 & 0.016663 & 0.001352 & 0.016663 & 3.242E-16 & 1.000 \\
C(010) & 0.054542 & 0.392766 & 12.304719 & 0.016714 & 0.017875 & 0.001280 & 0.017875 & 4.378E-16 & 1.000 \\
C(001) & 0.055179 & 0.392928 & 12.066565 & 0.016829 & 0.017636 & 0.000857 & 0.017636 & 3.363E-16 & 1.000 \\
C(110) & 0.056276 & 0.400342 & 14.090362 & 0.020401 & 0.021877 & 0.001575 & 0.021877 & 7.703E-16 & 1.000 \\
C(101) & 0.058387 & 0.436926 & 11.772486 & 0.020514 & 0.021813 & 0.001774 & 0.021813 & 7.490E-16 & 1.000 \\
C(011) & 0.057610 & 0.416929 & 14.085164 & 0.022258 & 0.023973 & 0.001995 & 0.023973 & 6.459E-16 & 1.000 \\
C(111) & 0.061990 & 0.490502 & 12.407313 & 0.027156 & 0.028870 & 0.001825 & 0.028870 & 1.370E-15 & 1.000 \\
\addlinespace[1.0pt]
\midrule
\addlinespace[0.4pt]
\multicolumn{10}{@{}l}{\textbf{hWOA}}\\[-0.8pt]
\addlinespace[0.2pt]
C(000) & 0.053121 & 0.392161 & 9.477940 & 0.012702 & 0.013650 & 0.001252 & 0.013650 & 2.602E-16 & 1.000 \\
C(100) & 0.053527 & 0.368177 & 12.634312 & 0.015438 & 0.016640 & 0.001448 & 0.016640 & 4.729E-16 & 1.000 \\
C(010) & 0.054499 & 0.391700 & 12.366443 & 0.016714 & 0.017840 & 0.001153 & 0.017840 & 4.003E-16 & 1.000 \\
C(001) & 0.055435 & 0.399276 & 11.715038 & 0.016828 & 0.018045 & 0.001753 & 0.018045 & 4.580E-16 & 1.000 \\
C(110) & 0.056987 & 0.417995 & 13.017109 & 0.020385 & 0.021374 & 0.001472 & 0.021374 & 1.368E-15 & 1.000 \\
C(101) & 0.057860 & 0.423489 & 12.468325 & 0.020513 & 0.021607 & 0.001269 & 0.021607 & 6.574E-16 & 1.000 \\
C(011) & 0.057947 & 0.425352 & 13.577249 & 0.022249 & 0.023505 & 0.001364 & 0.023505 & 8.283E-16 & 1.000 \\
C(111) & 0.061591 & 0.479859 & 12.913120 & 0.027147 & 0.028890 & 0.002125 & 0.028890 & 3.613E-15 & 1.000 \\
\addlinespace[1.0pt]
\midrule
\addlinespace[0.4pt]
\multicolumn{10}{@{}l}{\textbf{EO} \cite{AFeo2020}}\\[-0.8pt]
\addlinespace[0.2pt]
$C(000)$ & 0.051515 & 0.352537 & 11.538348 & 0.012666 & 0.012994 & 0.000581 & 0.012994 & 1.017E-16 & 1.000 \\
$C(100)$ & 0.054539 & 0.392920 & 11.202989 & 0.015431 & 0.016123 & 0.001059 & 0.016123 & 1.254E-16 & 1.000 \\
$C(010)$ & 0.054480 & 0.391215 & 12.394378 & 0.016714 & 0.017803 & 0.002104 & 0.017803 & 2.355E-16 & 1.000 \\
$C(001)$ & 0.054951 & 0.387330 & 12.390735 & 0.016831 & 0.017364 & 0.001097 & 0.017364 & 1.466E-16 & 1.000 \\
$C(110)$ & 0.057210 & 0.423621 & 12.702059 & 0.020385 & 0.020696 & 0.000320 & 0.020696 & 3.680E-16 & 1.000 \\
$C(101)$ & 0.057907 & 0.424667 & 12.404688 & 0.020512 & 0.020940 & 0.000529 & 0.020940 & 3.464E-16 & 1.000 \\
$C(011)$ & 0.058144 & 0.430328 & 13.290758 & 0.022245 & 0.022503 & 0.000228 & 0.022503 & 6.082E-16 & 1.000 \\
$C(111)$ & 0.061431 & 0.475641 & 13.122689 & 0.027145 & 0.027639 & 0.000494 & 0.027639 & 1.089E-14 & 1.000 \\
\addlinespace[1.0pt]
\midrule
\addlinespace[0.4pt]
\multicolumn{10}{@{}l}{\textbf{hEO}}\\[-0.8pt]
\addlinespace[0.2pt]
C(000) & 0.050833 & 0.336467 & 12.583033 & 0.012679 & 0.012981 & 0.000454 & 0.012981 & 8.344E-17 & 1.000 \\
C(100) & 0.054188 & 0.384238 & 11.674626 & 0.015429 & 0.015766 & 0.000356 & 0.015766 & 1.164E-16 & 1.000 \\
C(010) & 0.054339 & 0.387737 & 12.599449 & 0.016714 & 0.017230 & 0.000581 & 0.017230 & 2.916E-16 & 1.000 \\
C(001) & 0.055268 & 0.395119 & 11.943382 & 0.016828 & 0.017141 & 0.000395 & 0.017141 & 2.395E-16 & 1.000 \\
C(110) & 0.057312 & 0.426211 & 12.561060 & 0.020385 & 0.020888 & 0.000768 & 0.020888 & 4.631E-16 & 1.000 \\
C(101) & 0.057430 & 0.412710 & 13.074980 & 0.020520 & 0.020974 & 0.000552 & 0.020974 & 4.259E-16 & 1.000 \\
C(011) & 0.058657 & 0.443451 & 12.579895 & 0.022246 & 0.023039 & 0.000745 & 0.023039 & 5.963E-16 & 1.000 \\
C(111) & 0.061188 & 0.469265 & 13.449826 & 0.027144 & 0.027714 & 0.000870 & 0.027714 & 7.854E-16 & 1.000 \\
\addlinespace[1.0pt]
\midrule
\addlinespace[0.4pt]
\multicolumn{10}{@{}l}{\textbf{GA} \cite{KD1995},\cite{KD2000} }\\[-0.8pt]
\addlinespace[0.2pt]
$C(000)$ & 0.052761 & 0.383063 & 9.896722 & 0.012686 & 0.014564 & 0.001777 & 0.014564 & 1.835E-17 & 1.000 \\*
$C(100)$ & 0.053969 & 0.378857 & 11.983974 & 0.015431 & 0.017348 & 0.002104 & 0.017348 & 8.918E-18 & 1.000 \\*
$C(010)$ & 0.053904 & 0.377142 & 13.259274 & 0.016722 & 0.019681 & 0.002639 & 0.019681 & 1.763E-17 & 1.000 \\*
$C(001)$ & 0.055970 & 0.412733 & 11.021647 & 0.016836 & 0.018958 & 0.002429 & 0.018958 & 8.843E-18 & 1.000 \\*
$C(110)$ & 0.059302 & 0.478682 & 10.163892 & 0.020477 & 0.022618 & 0.002237 & 0.022618 & 1.270E-17 & 1.000 \\*
$C(101)$ & 0.058202 & 0.432164 & 12.011956 & 0.020513 & 0.022761 & 0.002456 & 0.022761 & 7.550E-15 & 1.000 \\*
$C(011)$ & 0.057856 & 0.423066 & 13.711844 & 0.022250 & 0.024026 & 0.001859 & 0.024026 & 7.367E-17 & 1.000 \\*
$C(111)$ & 0.062127 & 0.494177 & 12.241819 & 0.027165 & 0.030266 & 0.002656 & 0.030266 & 7.005E-17 & 1.000 \\
\addlinespace[1.0pt]
\midrule
\addlinespace[0.4pt]
\multicolumn{10}{@{}l}{\textbf{hGA}}\\[-0.8pt]
\addlinespace[0.2pt]
$C(000)$ & 0.051408 & 0.349984 & 11.695948 & 0.012668 & 0.013524 & 0.001391 & 0.013524 & 3.193E-18 & 1.000 \\*
$C(100)$ & 0.054232 & 0.385302 & 11.615348 & 0.015429 & 0.017746 & 0.002294 & 0.017746 & 7.270E-18 & 1.000 \\*
$C(010)$ & 0.054542 & 0.392750 & 12.305667 & 0.016714 & 0.019236 & 0.002559 & 0.019236 & 6.249E-18 & 1.000 \\*
$C(001)$ & 0.055624 & 0.403982 & 11.465529 & 0.016831 & 0.018878 & 0.002259 & 0.018878 & 5.893E-18 & 1.000 \\*
$C(110)$ & 0.057934 & 0.442214 & 11.745103 & 0.020401 & 0.022012 & 0.002282 & 0.022012 & 2.136E-17 & 1.000 \\*
$C(101)$ & 0.057556 & 0.415839 & 12.894134 & 0.020517 & 0.021866 & 0.001483 & 0.021866 & 2.138E-17 & 1.000 \\*
$C(011)$ & 0.058293 & 0.434091 & 13.081225 & 0.022246 & 0.024722 & 0.002556 & 0.024722 & 1.878E-17 & 1.000 \\*
$C(111)$ & 0.061467 & 0.476575 & 13.075816 & 0.027145 & 0.028474 & 0.001677 & 0.028474 & 1.758E-17 & 1.000 \\
\bottomrule
\end{tabular}
}
}
\par\vspace{3pt}
\begin{minipage}{\textwidth}
\footnotesize

\end{minipage}
\endgroup
\end{table*}
\clearpage
The symbols in Table~\ref{tab:MPR_all_results} (p.13-14) are interpreted as follows: $x_1$, $x_2$ and $x_3$ denote the \texttt{spring geometry parameters} that yield $F_{best}$ - the best penalized objective value. $\overline{F}$ and $SD(F)$ are its mean and standard deviation;
$\overline{f}$ is the mean unpenalized objective value;
$\overline{V}$ is the mean constraint violation measure;
and FR is the feasibility rate.

For each algorithm, $C(000)$ produced the lowest $F_{best}$, while $C(111)$ produced the highest. Thirteen algorithms achieved an $FR$ of 1.000 in all eight configurations. The exceptions were CMA-ES and DE. CMA-ES had an FR of less than 1 for $C(010)$, $C(110)$, $C(011)$ and $C(111)$. 
DE yielded $FR = 0.933$ for $C(011)$ and $C(111)$. L-SHADE, hL-SHADE, MPA and hMPA maintained SD values of no more than $3\times10^{-6}$ across all configurations.


\subsection{Adapted performance space assessment of the spring Constrained Single-Objective Optimization Problem (CSOP)}

The performance-space framework of Vodopija et al. in \cite{AV2025} was adapted to the eight $MPR$ configurations to assess their ability to differentiate algorithm behaviour 
throughout the optimisation process, rather than solely ranking the algorithms by their 
final outcomes. This preserves the original benchmark characterisation objective of the framework 
while extending its scope from Constrained Multiobjective Optimization Problem (CMOP, e.g., Filipi{\v{c}} and  Vodopija in \cite{BF2021}) test suites to distinct configurations of a constrained, single objective engineering problem.

As the $MPR$ spring benchmark is a  
CSOP (e.g., Section \ref{RW}), no Pareto front is defined. 
Therefore, the Empirical Runtime Distribution (ERD) and algorithm differentiation principles are retained,
whereas the Pareto-front approximation branch is replaced by a
CSOP-specific quality branch. Each $MPR$ configuration is treated as a
separate performance-space instance, and runtime observations are not
pooled across configurations.

For CSOPs, Vodopija et al. in \cite{AV2025} report the  COmparing Continuous Optimizers (COCO) indicator based
on the best-so-far value of $f+v$. In the present adaptation, $v$ is
replaced by the quadratic worst-case $MPR$ constraint-violation measure
$V$ defined in Eq.~\eqref{eq:Viol}, thereby preserving consistency with
the engineering formulation. For configuration $C$, the logged best-so-far 
indicator after $FE$ function evaluations is as follows:
\begin{equation}
I(C, FE)
=
\min_{1\leq t\leq FE}
\left[
f\left(x^{(t)}\right)+V\left(x^{(t)}; M,P,R\right)
\right],
\label{eq:IC}
\end{equation}
where $x^{(t)}$ denotes the candidate design evaluated at function evaluation $t$.

Equation~\eqref{eq:IC} deliberately retains the violation measure of the
engineering model rather than the linear aggregation used by
Vodopija et al. in \cite{AV2025}, who note that the definition of overall
constraint violation can affect the analysis. The logger does not alter
the optimisation process. All algorithms continue to minimise the
penalised objective in Eq.~\eqref{eq:MPR_penalty}; hence, the penalty
coefficient $10^6$ is not included in $I(C, FE)$. The best-so-far violation
and the best feasible objective value are stored separately.

Analytically established reference values are not available for the
proposed $MPR$ configurations. Therefore, the reference used for target
construction is defined empirically as follows:
\begin{equation}
I_{\mathrm{ref}}(C)
=
\min_{A,r}
I(C, FE_{\max}, A, r),
\label{eq:Iref}
\end{equation}
where $A$ and $r$ denote the algorithm and (independent) run, respectively, and
$FE_{\max}$ is the maximum evaluation budget.

Thus, as defined in Eq.~\eqref{eq:Iref}, $I_{\mathrm{ref}}(C)$ represents
the best observed CSOP indicator value for configuration $C$ across all
15 algorithms and 30 runs. It is an empirical benchmark within
this study, not a claim of global optimality. Vodopija et al. in~\cite{AV2025}
show that the reference used for target construction can be based on an
estimated value. Benchmark references based on the best value attained by the considered
solvers have also been used in optimisation benchmarking
(e.g., Mor\'e and Wild~\cite{JJM2009},
Ma et al.~\cite{KM2025}, and Zaryab et al.~\cite{SAZ2025}).

The target precisions and order-of-magnitude scaling follow the procedure
of Vodopija et al. in \cite{AV2025}, with the 51 target precisions defined
as follows: $\epsilon \in\{10^{-5},10^{-4.9},\ldots,10^{-0.1},$ $10^{0}\}$.
The magnitudes of constraint violations differ among the $MPR$ configurations and are therefore normalised separately. For each configuration, 100 designs are sampled uniformly from the decision domain. The violation is scaled by the order of magnitude of the corresponding sample median: $10^{\lfloor\log_{10}(\operatorname{median} V)\rfloor}$, where 
$\lfloor\cdot\rfloor$ denotes the floor function.
If the sampled median is zero, a scaling factor of one is used. 

The two adapted target families are then defined as:
\begin{equation}
\tau^{+}(C)=I_{\mathrm{ref}}(C)+\epsilon,
\qquad
\tau^{-}=\epsilon .
\label{eq:targets}
\end{equation}
The two branches in Eq.~\eqref{eq:targets} have distinct roles.
The $\tau^{+}$ branch measures progress in the composite CSOP indicator
$f+V$ towards the empirical reference. It does not represent Pareto-front
approximation or a feasibility-gated objective measure.
The $\tau^{-}$ branch measures progress in reducing the normalised
constraint violation, $V$. For each run and target, runtime is
defined as the first function evaluation at which the corresponding
threshold is reached.

For each algorithm, the ERD aggregates the 30 independent runs and two
sets of 51 targets. Each curve therefore represents $30\times102$
target-runtime observations and shows the proportion of targets reached
within the evaluation budget. Runtime is displayed as
$\log_{10}(FE/dim)$ (here $dim=3$), following the dimension-normalised logarithmic
representation of Vodopija et al. in \cite{AV2025}.

Algorithm differentiation is computed separately for the two target branches.
An unreached target is assigned $FE_{\max}$. Let $T_{(k)}^{\pm}(A)$ denote 
the $k$-th ordered runtime for algorithm $A$
after aggregation over the 30 runs and 51 targets of a given branch (and analogously for \(B\)).
With $N=30\times51$, the separation is defined as follows:
\begin{equation}
\Delta^{\pm}(A,B)
=
\frac{1}{N\cdot\log_{10}(FE_{\max})}
\sum_{k=1}^{N}
\left|
\log_{10}\left(T_{(k)}^{\pm}(A)\right)
-
\log_{10}\left(T_{(k)}^{\pm}(B)\right)
\right|.
\label{eq:delta}
\end{equation}
Equation~\eqref{eq:delta} represents the normalised area between the
corresponding aggregated ERDs in logarithmic runtime space.
For each algorithm pair and MPR configuration, one $\Delta^{+}$ and one
$\Delta^{-}$ are obtained.
The $\Delta^{+}$ measure quantifies differentiation in attaining the
adapted CSOP quality targets.
The $\Delta^{-}$ measure quantifies differentiation in constraint
satisfaction. As both branches contain 51 targets, they receive equal weight:
\begin{equation}
\Delta(A,B)
=
\frac{1}{2}
\left[
\Delta^{-}(A,B)+\Delta^{+}(A,B)
\right].
\label{eq:delta_total}
\end{equation}

Following Vodopija et al. in \cite{AV2025}, 
Eq.~\eqref{eq:delta_total} represents the equal-weight form of the
decomposition.

Small values of $\Delta$ indicate similar runtime behaviour.
Larger values indicate stronger performance-space separation.
With 15 algorithms, each $MPR$ configuration yields
$\binom{15}{2}=105$ algorithm pairs.
Thus, 105 pairwise values are obtained for each of
$\Delta^{+}$, $\Delta^{-}$, and $\Delta$.

The resulting ERDs and pairwise differentiation distributions for the eight MPR 
configurations analysed independently are presented in Fig.~\ref{fig2}.

\begin{figure}[ht]
\centering
    \centering
      \noindent\textit{Figure~\ref{fig2} (continued).}\par\vspace{20pt}
    \includegraphics[
        width=\linewidth,
        keepaspectratio
    ]{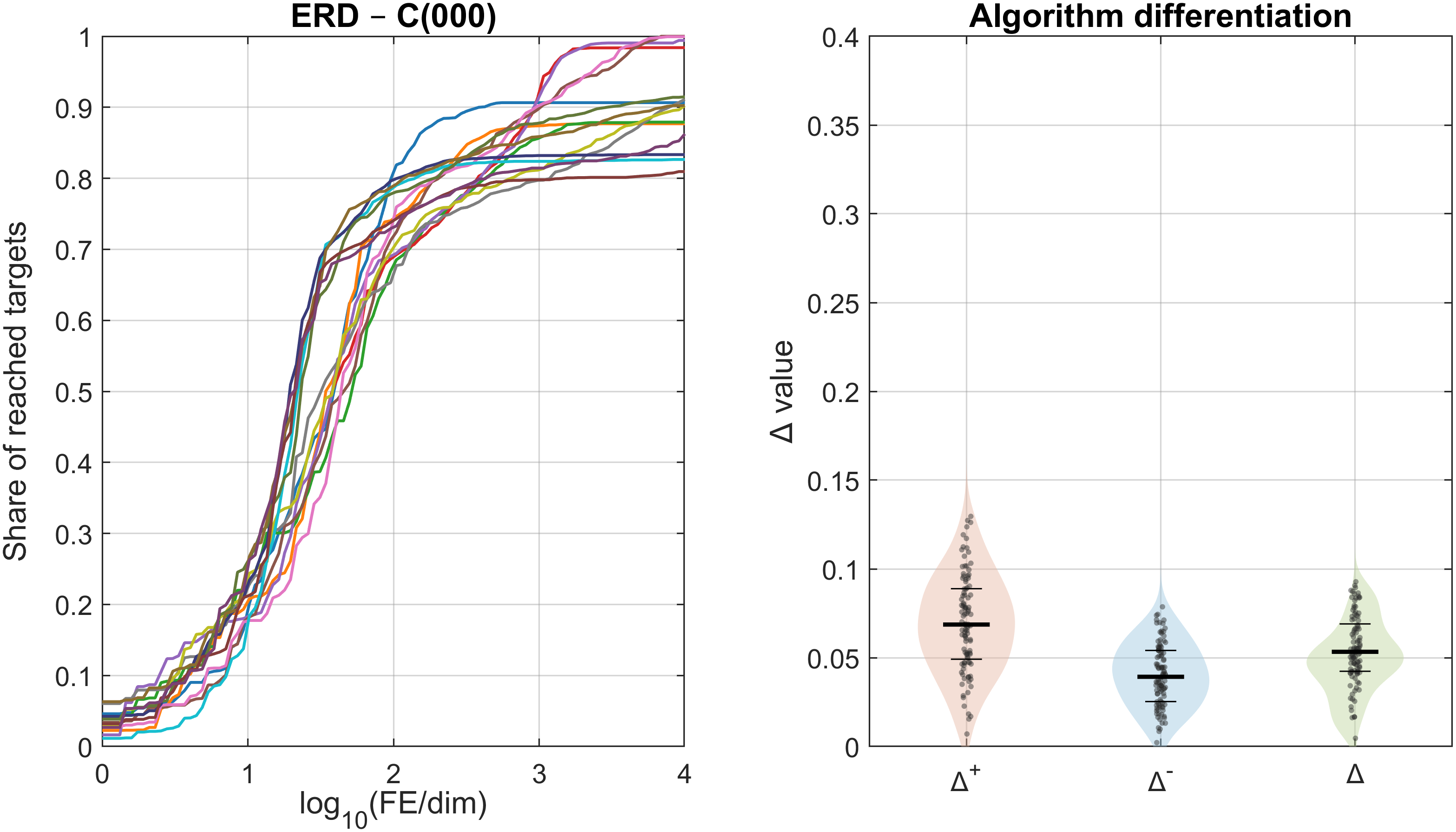}
\hfill
\vspace{0.9cm}
    \centering
    \includegraphics[
        width=\linewidth,
        keepaspectratio
    ]{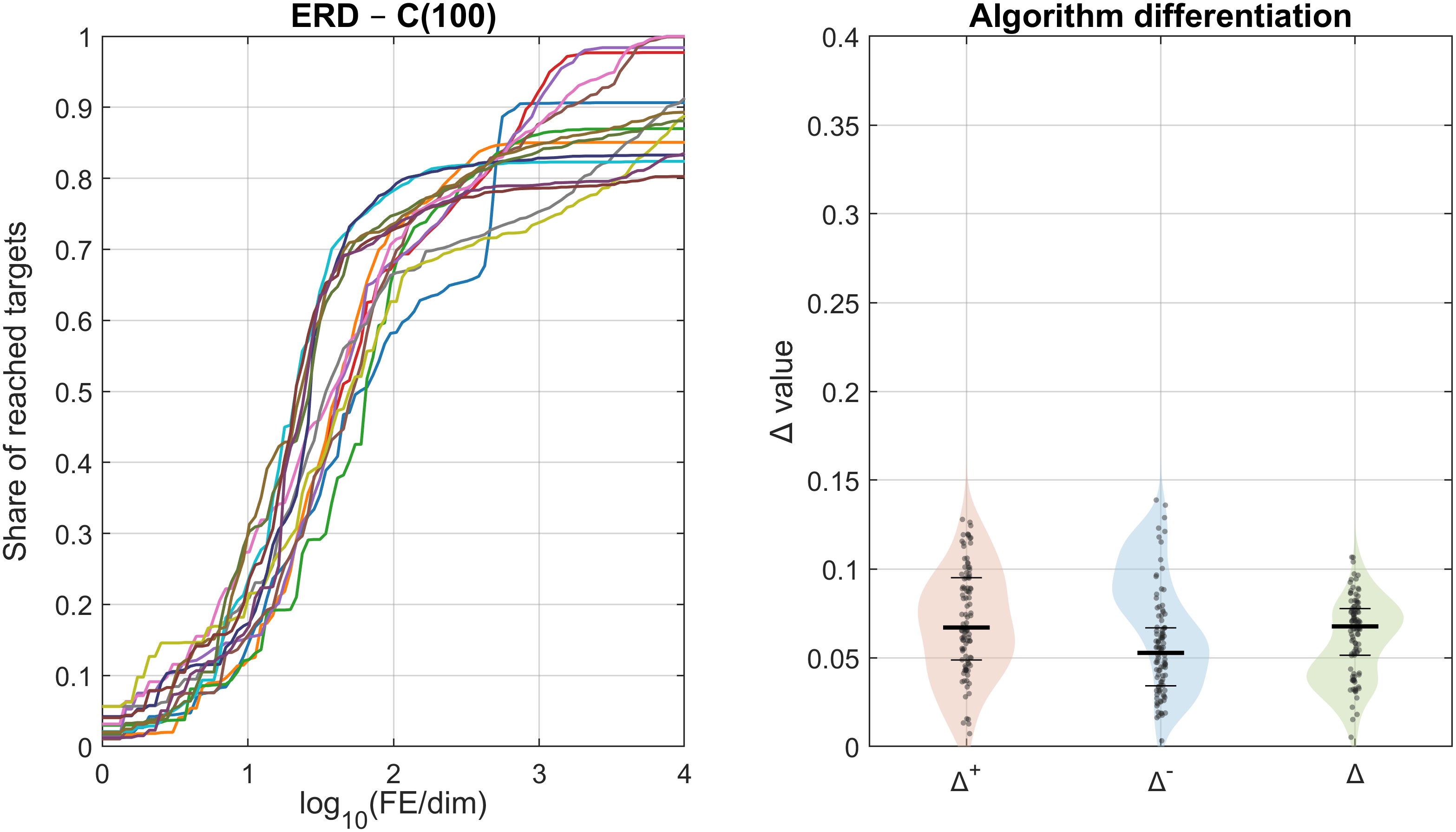}
\end{figure}

\clearpage

\begin{figure}[ht]
\centering
\begin{subfigure}{1.0\textwidth}
    \centering
    \noindent\textit{Figure~\ref{fig2} (continued).}\par\vspace{20pt}
    \includegraphics[
        width=\linewidth,
        keepaspectratio
    ]{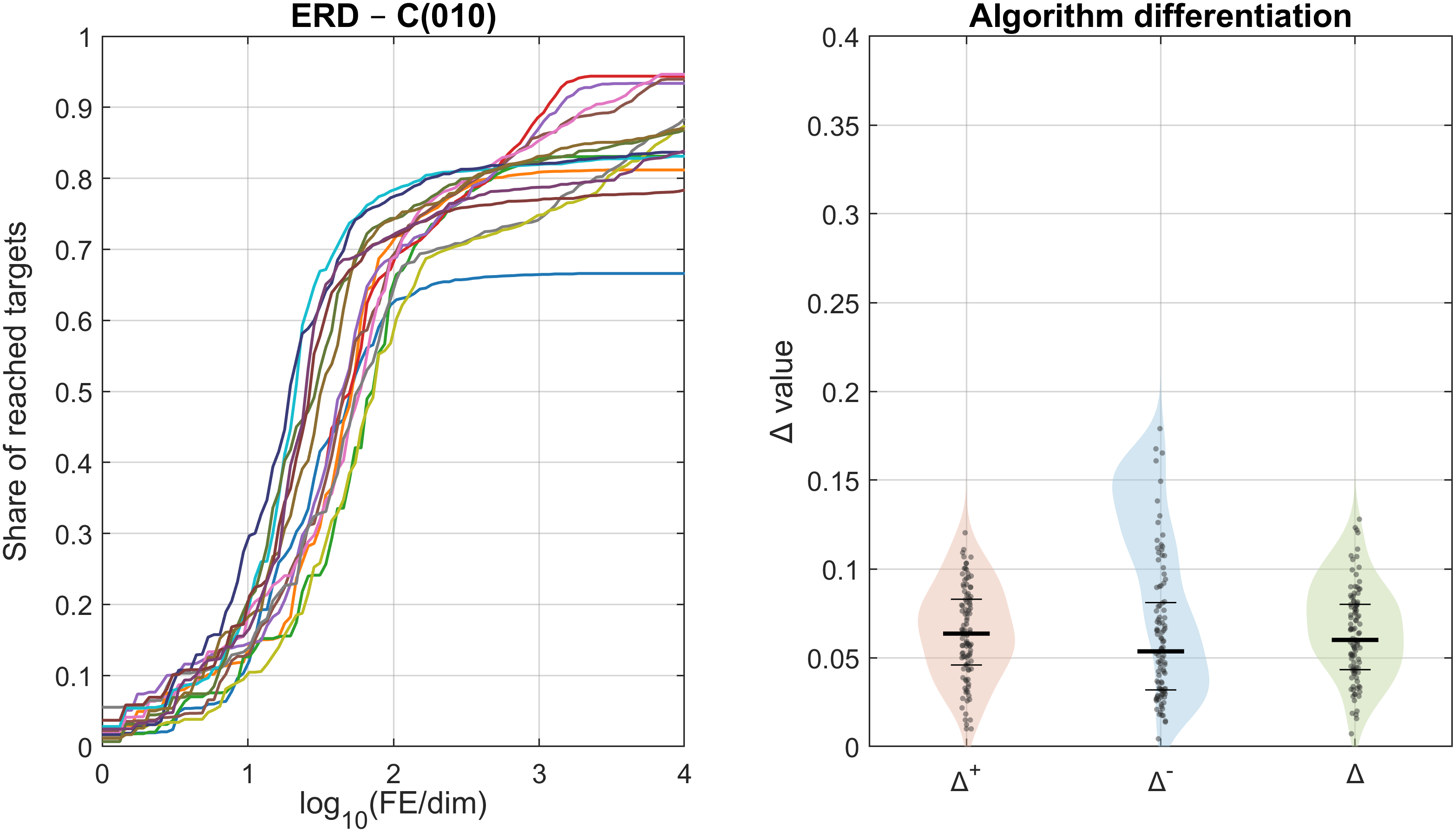}
\end{subfigure}
\hfill
\vspace{0.9cm}
\begin{subfigure}{1.0\textwidth}
    \centering
    \includegraphics[
        width=\linewidth,
        keepaspectratio
    ]{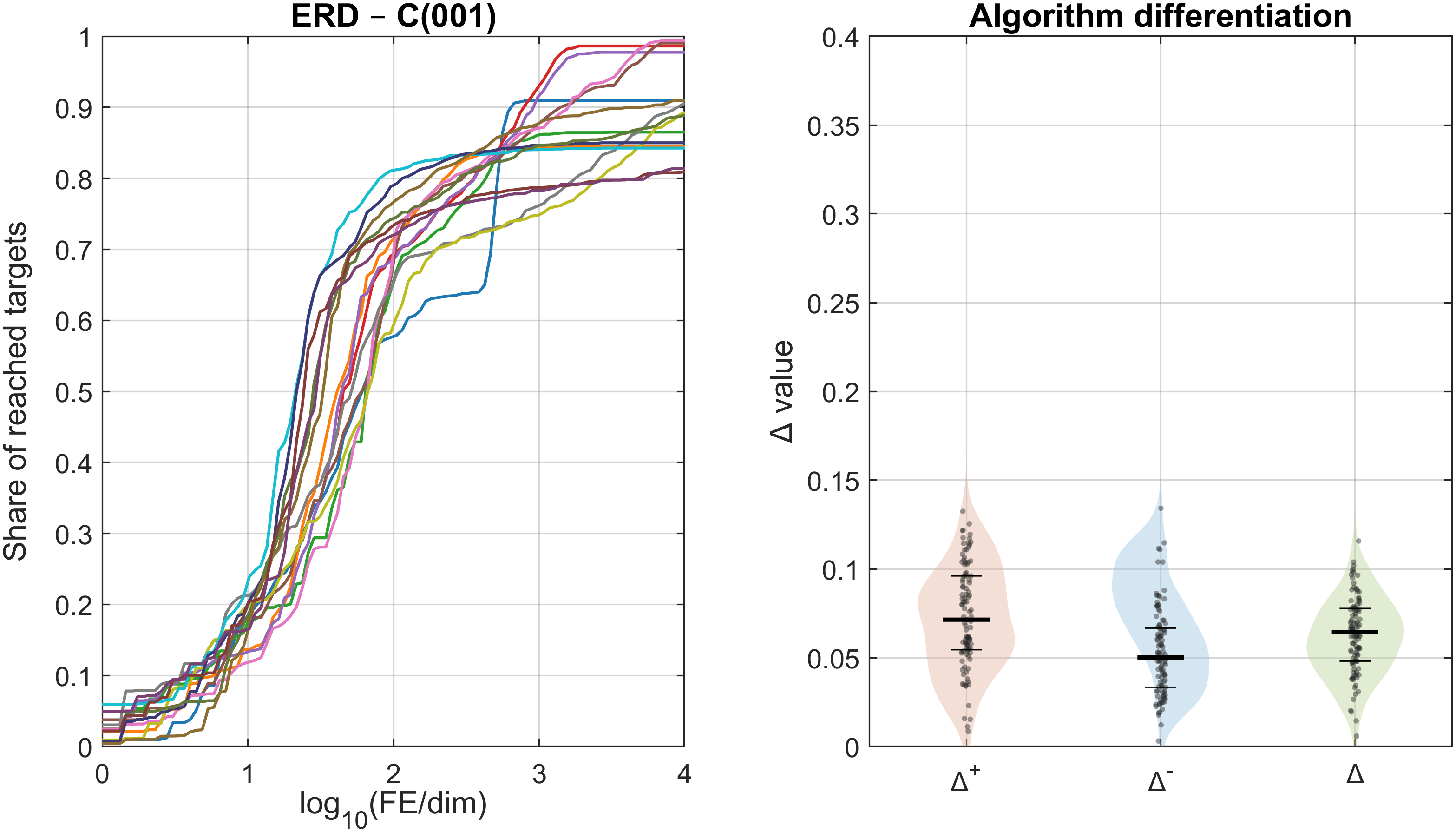}
\end{subfigure}
\end{figure}

\begin{figure}[ht]
\ContinuedFloat
\centering
\begin{subfigure}{1.0\textwidth}
    \centering
     \noindent\textit{Figure~\ref{fig2} (continued).}\par\vspace{20pt}
    \includegraphics[
        width=\linewidth,
        keepaspectratio
    ]{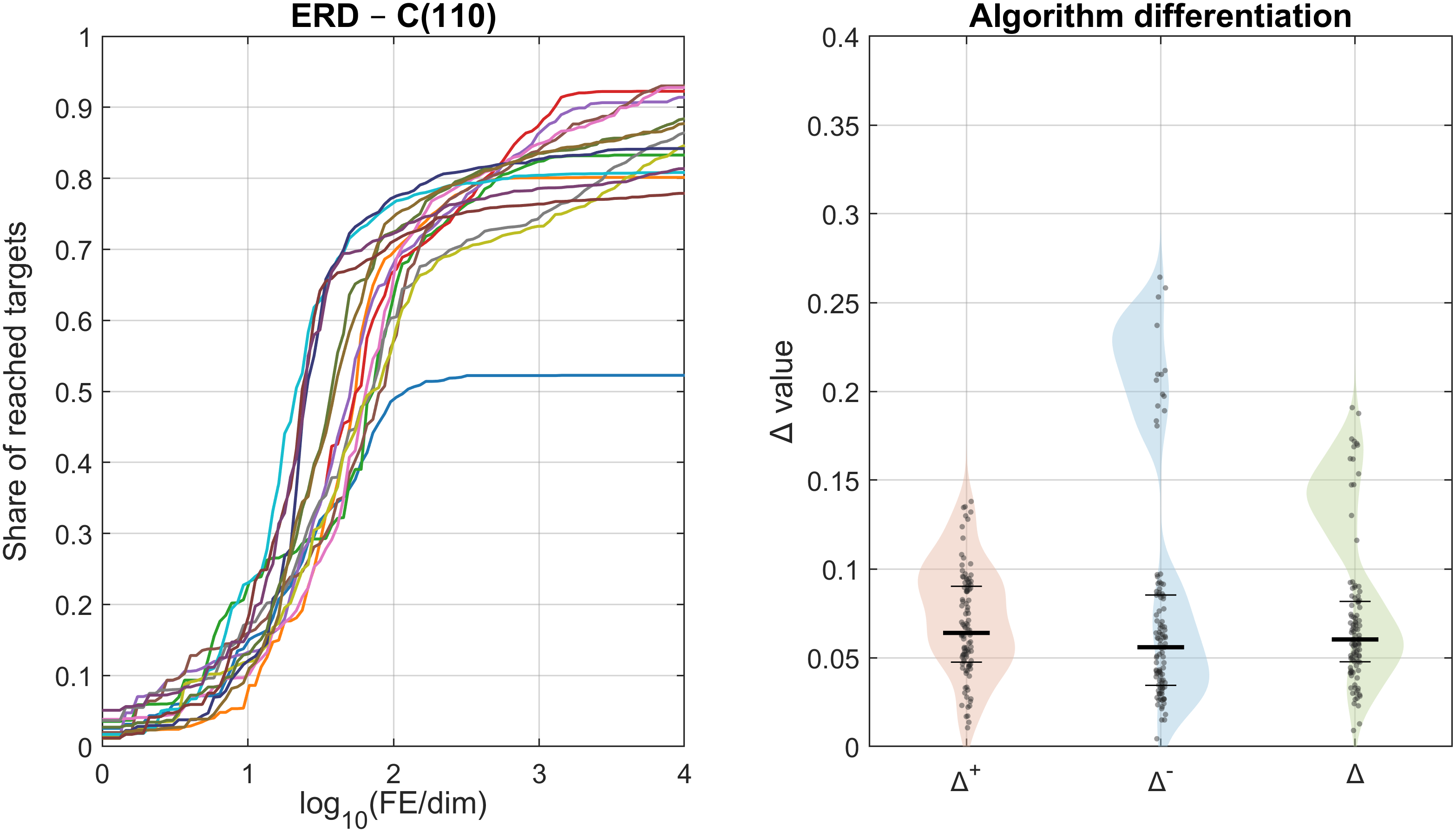}
\end{subfigure}
\hfill
\vspace{0.9cm}
\begin{subfigure}{1.0\textwidth}
    \centering
    \includegraphics[
        width=\linewidth,
        keepaspectratio
    ]{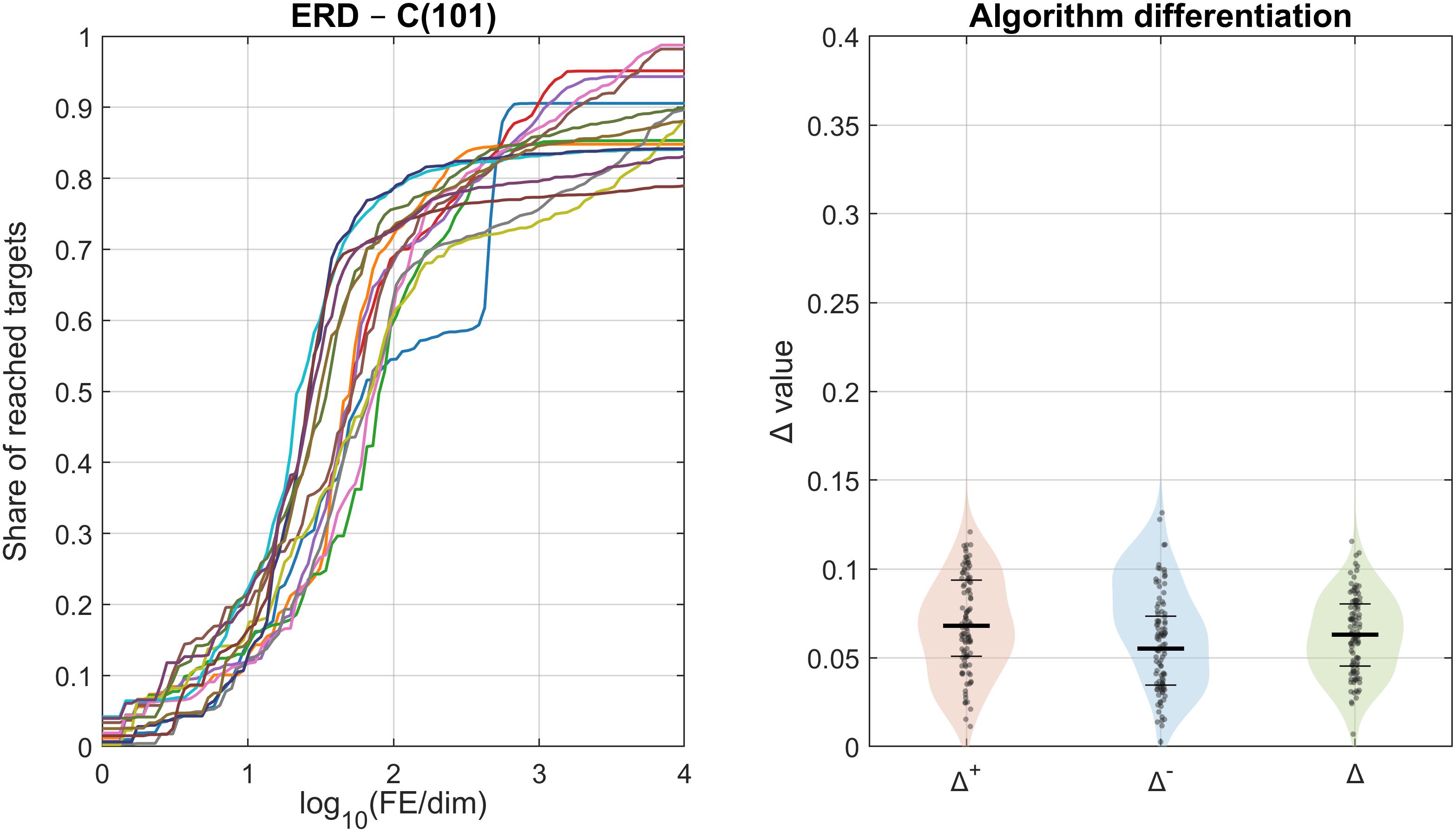}
\end{subfigure}
\end{figure}

\clearpage

\begin{figure}[ht]
\ContinuedFloat
\centering
\begin{subfigure}{1.0\textwidth}
    \centering
     \noindent\textit{Figure~\ref{fig2} (continued).}\par\vspace{20pt}
    \includegraphics[
        width=\linewidth,
        keepaspectratio
    ]{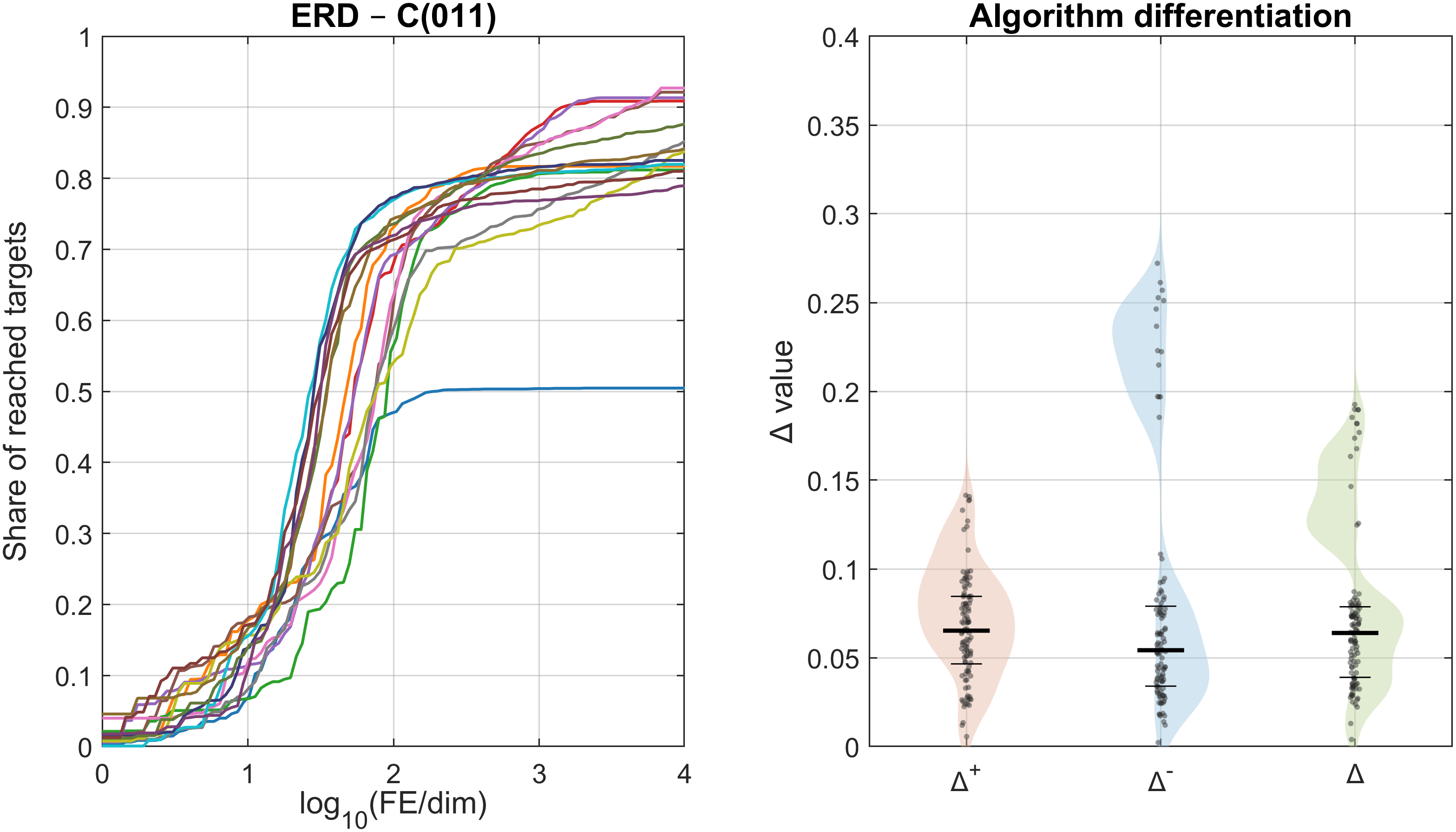}
\end{subfigure}
\hfill
\vspace{0.9cm}
\begin{subfigure}{1.0\textwidth}
    \centering
    \includegraphics[
        width=\linewidth,
        keepaspectratio
    ]{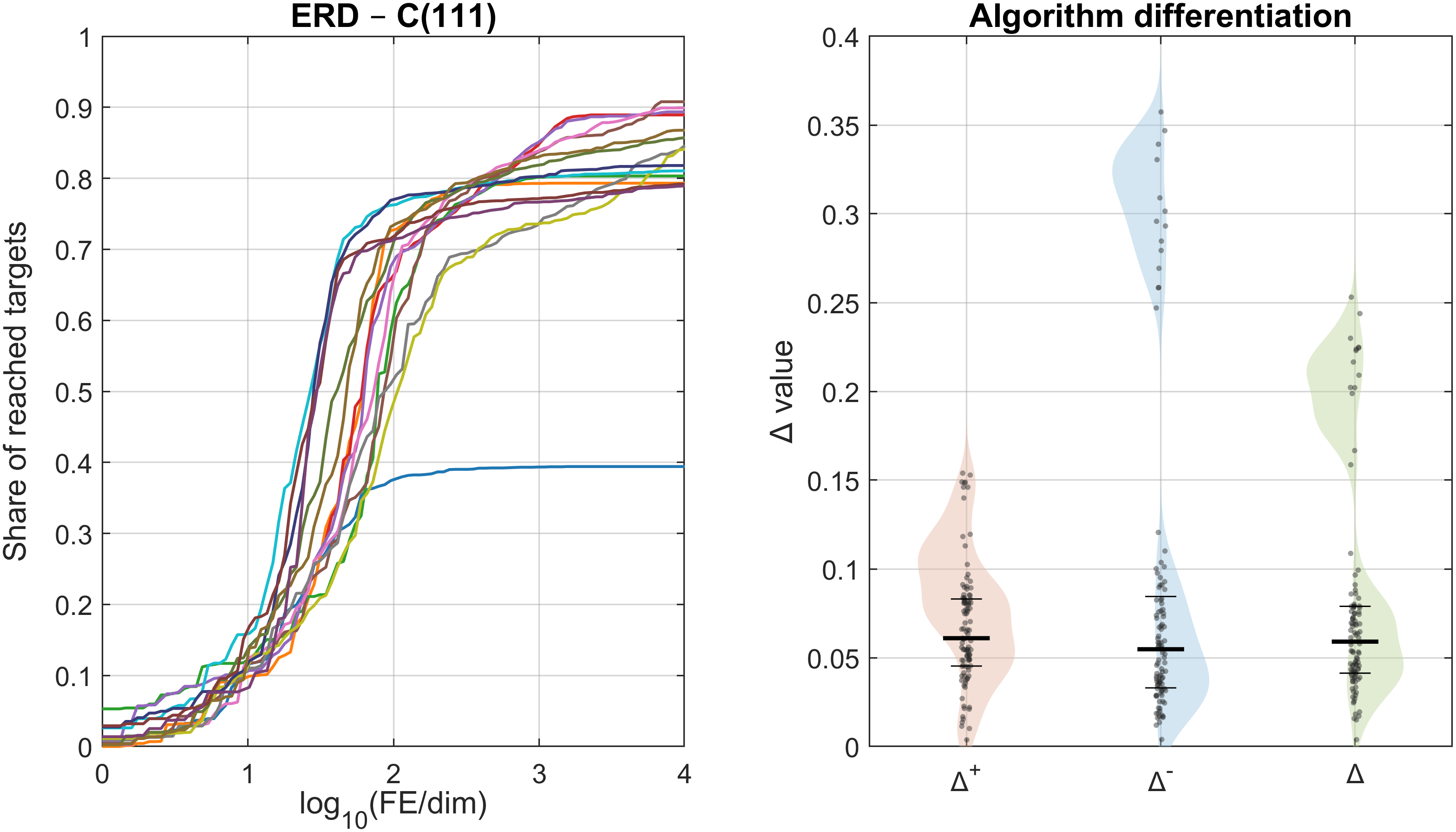}
\end{subfigure}
 \centering
\vspace{-0.39cm}
\begin{subfigure}{1.08\textwidth}
    \centering
    \includegraphics[
        width=\linewidth,
        keepaspectratio
    ]{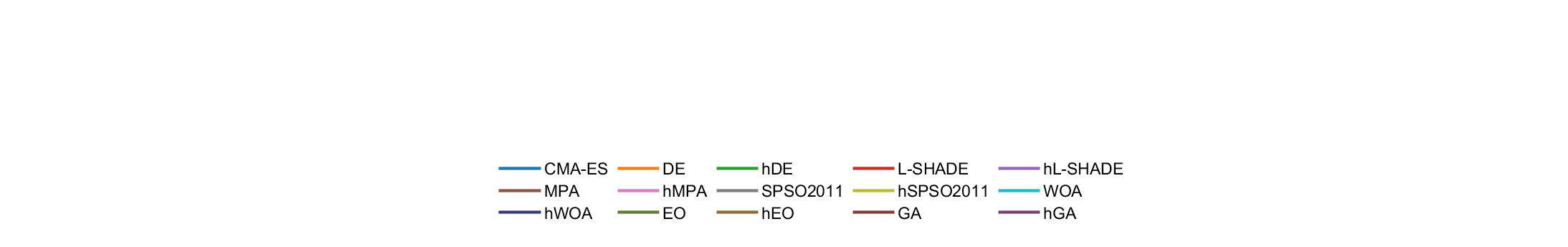}
\end{subfigure}
\vspace{0.49cm}
\begin{subfigure}{0.54\textwidth}
\noindent\hspace*{0.1\textwidth}%
    \centering
    \includegraphics[
        width=\linewidth,
        keepaspectratio
    ]{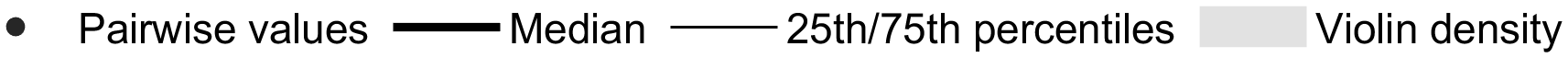}
\end{subfigure}
 \caption{Performance-space characterisation of the $MPR$ configurations.}
 \label{fig2}
\end{figure}
\clearpage
Figure~\ref{fig2} shows broadly similar ERD profiles across the $MPR$ configurations.
The degree of algorithm separation varies by configuration.
Overall, the median separation for $\Delta^{+}$ generally exceeds that for $\Delta^{-}$, with moderate pairwise differences. Pronounced upper tails occur for $\Delta^{-}$ in selected combined configurations, notably $C(011)$, $C(110)$, and $C(111)$. The aggregate $\Delta$ indicates that stronger separation is concentrated among specific algorithm pairs.

\subsection{Fixed target expected runtime analysis}

The preceding performance-space assessment establishes whether the $MPR$ configurations affect the behaviour of algorithms during optimisation. 
However, it does not directly measure the effort required to reach a prescribed
accuracy in the feasible unpenalised objective.
This distinction is important for engineering benchmarks. Solution quality must be considered alongside the effort required to achieve it under the imposed robustness conditions. The present analysis therefore provides a complementary view of the cost of reaching the target and configuration-wide coverage. To this end, the fixed-target expected-runtime framework of Opara and Arabas in~\cite{KO2011} has been adapted for the $MPR$ benchmark.

Opara and Arabas employed fixed target analysis to quantify the computational effort required to achieve a specified level of optimisation. They also demonstrate how runtime information can be aggregated across benchmark problems and accuracy levels. In this case, the benchmark-problem index is replaced by the $MPR$ configuration index. This is a deliberate adaptation. The eight configurations are not unrelated optimisation problems. Rather, they are systematically related variants of the same TCSD problem. They arise from the factorial activation of $M$, $P$, and $R$, as defined in Section~\ref{subsec:MPR}. The resulting analysis therefore measures computational efficiency across a controlled family of robustness conditions.

The literature reviewed for this study revealed no comparable joint application.
In particular, the two frameworks have not previously been used together for
systematically related configurations of a single constrained engineering benchmark.
The present work therefore addresses this gap.

The two assessments use different performance criteria. The earlier ERD
construction, defined in Eqs.~(\ref{eq:IC})-(\ref{eq:delta_total}), combines
the composite CSOP indicator $f+V$ with separate constraint-violation targets.
In contrast, the fixed-target assessment considers only feasible solutions
$V=0$ and evaluates them using the unpenalised objective function $f$.

For algorithm $A$, configuration $C$, and independent run $r$, the best feasible
unpenalised objective value available after $FE$ function evaluations is denoted as follows:
$$f_{\mathrm{best}}(C, FE, A, r)
=\min_{\substack{1 \leq t \leq FE \\
V\!\left(x^{(t)};M,P,R\right)=0}}
f\!\left(x^{(t)}\right),
\quad~\text{for} \ \
1 \leq FE \leq FE_{\max},$$
where $t$ denotes the function evaluation index. If no feasible solution has been
found within the first $FE$ function evaluations, then
$f_{\mathrm{best}}(C, FE, A, r)=+\infty$.

A separate configuration specific reference is then defined as:
$f_{\mathrm{ref}}(C)=\min_{A,r}f_{\mathrm{best}}(C,$ $FE_{\max}, A, r),$
where the minimum is taken over all 15 algorithms and 30 independent runs for
configuration $C$. This reference is fixed within each configuration. As with the
empirical reference in Eq.~(\ref{eq:Iref}), $f_{\mathrm{ref}}(C)$ is the best-known
value within the predefined solver set and is not interpreted as an analytically
established global optimum. The comparison is therefore an explicit within-study
fixed-target assessment.

For an accuracy level of $\Delta f$ the $r$-th run of algorithm $A$ on configuration $C$ reaches the target at a function-evaluation count of FE when:
\begin{equation}
f_{\mathrm{best}}(C, FE, A, r)
-
f_{\mathrm{ref}}(C)
\leq
\Delta f.
\label{eq:ft_target}
\end{equation}
The runtime of a successful run is defined as the first function evaluation at which 
Eq.~\eqref{eq:ft_target} is satisfied.  Following Opara and Arabas~\cite{KO2011}, 
the estimated expected runtime is defined for algorithm $A$,
configuration $C$, and accuracy level $\Delta f$. This quantity is denoted by:
$\widehat{E}\!\left(RT(A, C, \Delta f)\right)$.

The estimate is obtained by dividing the total FE cost accumulated over all 30 runs 
by the number of successful runs. For successful runs, the time taken to hit the first 
target is used, whereas for unsuccessful runs, the FE cost incurred at termination is used. 
If no run reaches the target, then:
$\widehat{E}\!\left(RT(A, C, \Delta f)\right)$ is assigned $+\infty$.

The same grid of 51 logarithmically spaced precision levels, as defined in the
previous subsection, is used here. These levels specify the allowable gap in the
feasible unpenalised objective relative to $f_{\mathrm{ref}}(C)$. Four
representative levels are shown separately:
$\Delta f\in\{10^{-2},10^{-3},10^{-4},10^{-5}\}$.
The corresponding results are shown in Fig.~\ref{fig:target_ert}.

\begin{figure}[!htbp]
\centering
 \noindent\textit{Figure~\ref{fig:target_ert} (continued).}\par\vspace{15pt}
\includegraphics[
    width=0.9\textwidth,
    keepaspectratio
]{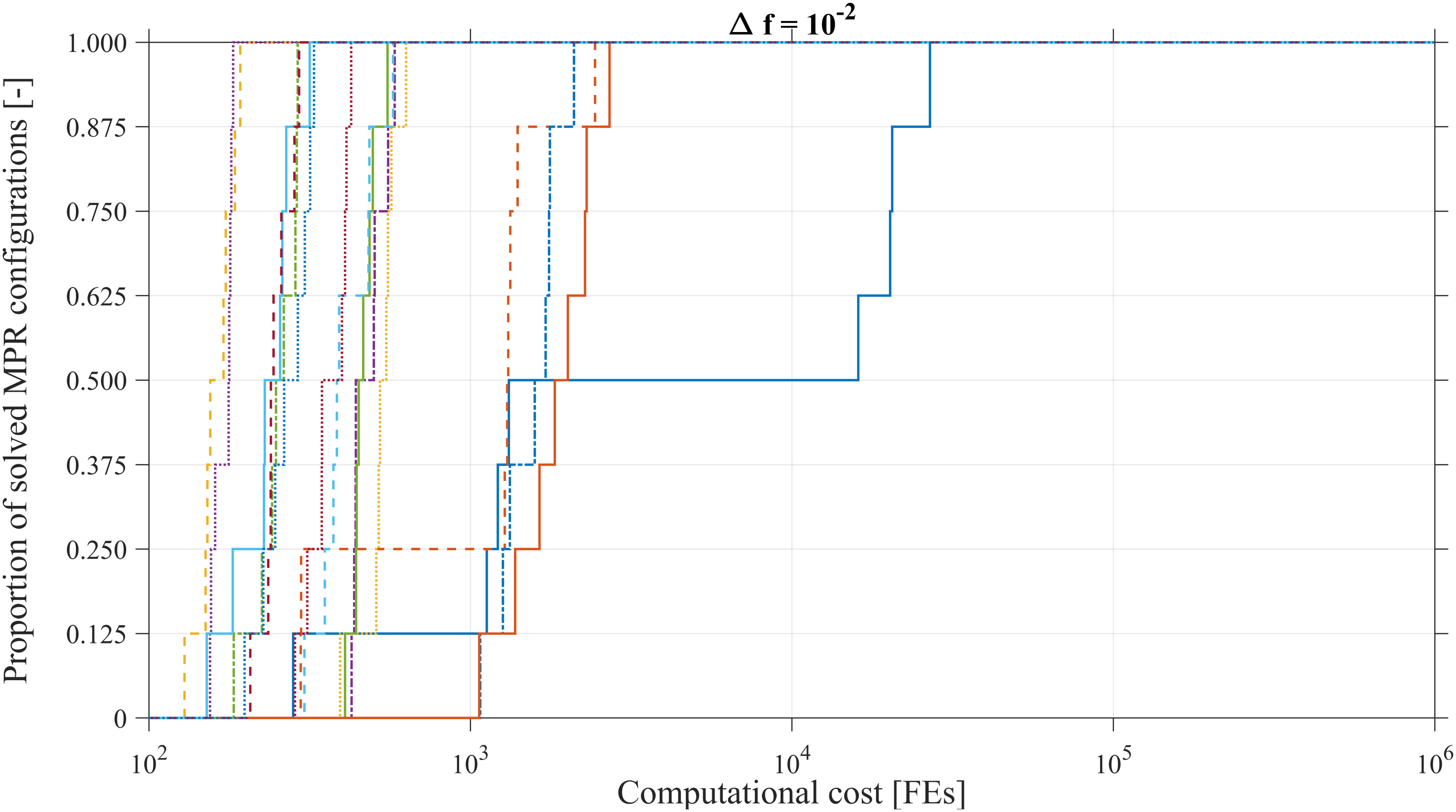}

\vspace{0.4cm}

\includegraphics[
    width=0.9\textwidth,
    keepaspectratio
]{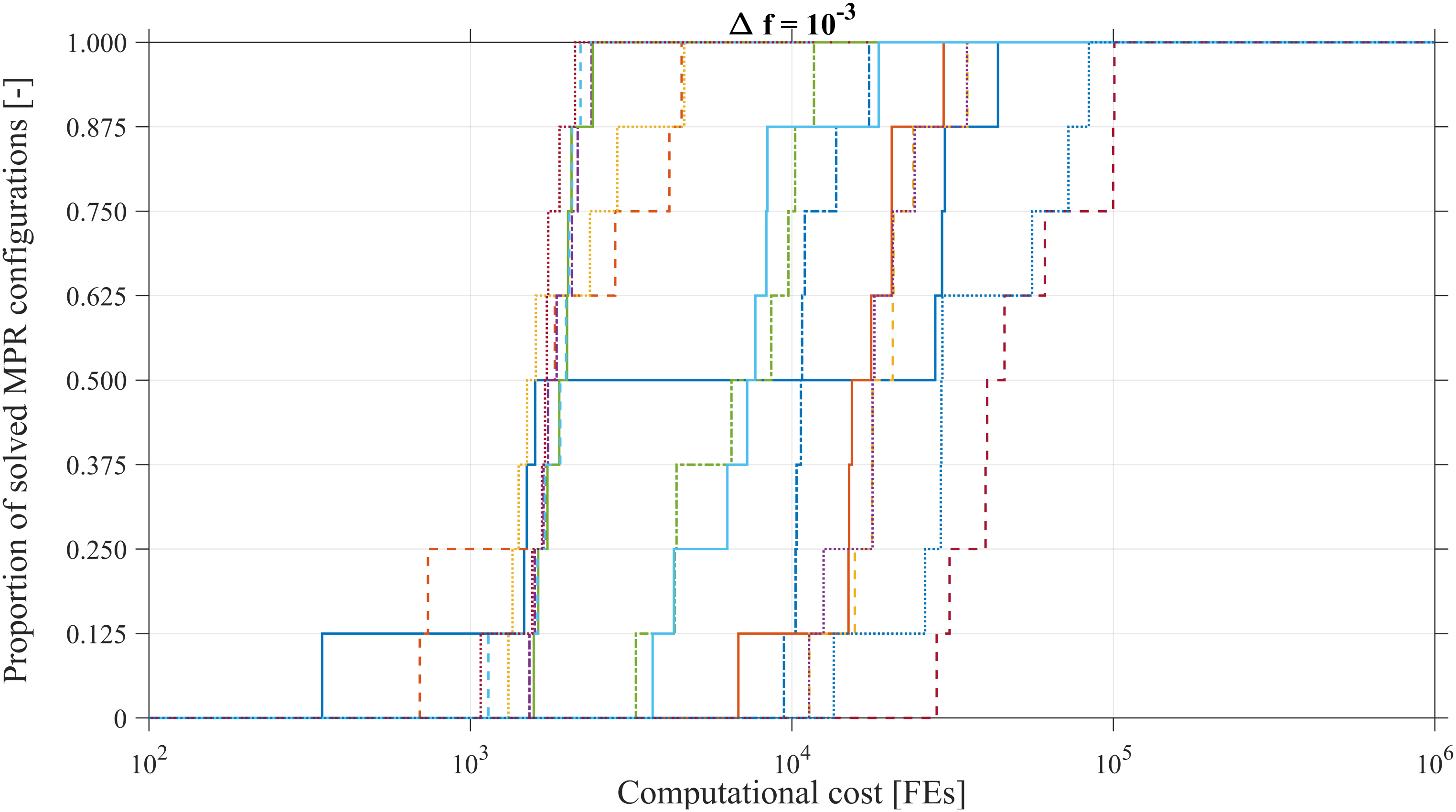}
\end{figure}

\clearpage

\begin{figure}[!htbp]
\centering
 \noindent\textit{Figure~\ref{fig:target_ert} (continued).}\par\vspace{15pt}
\includegraphics[
    width=0.9\textwidth,
    keepaspectratio
]{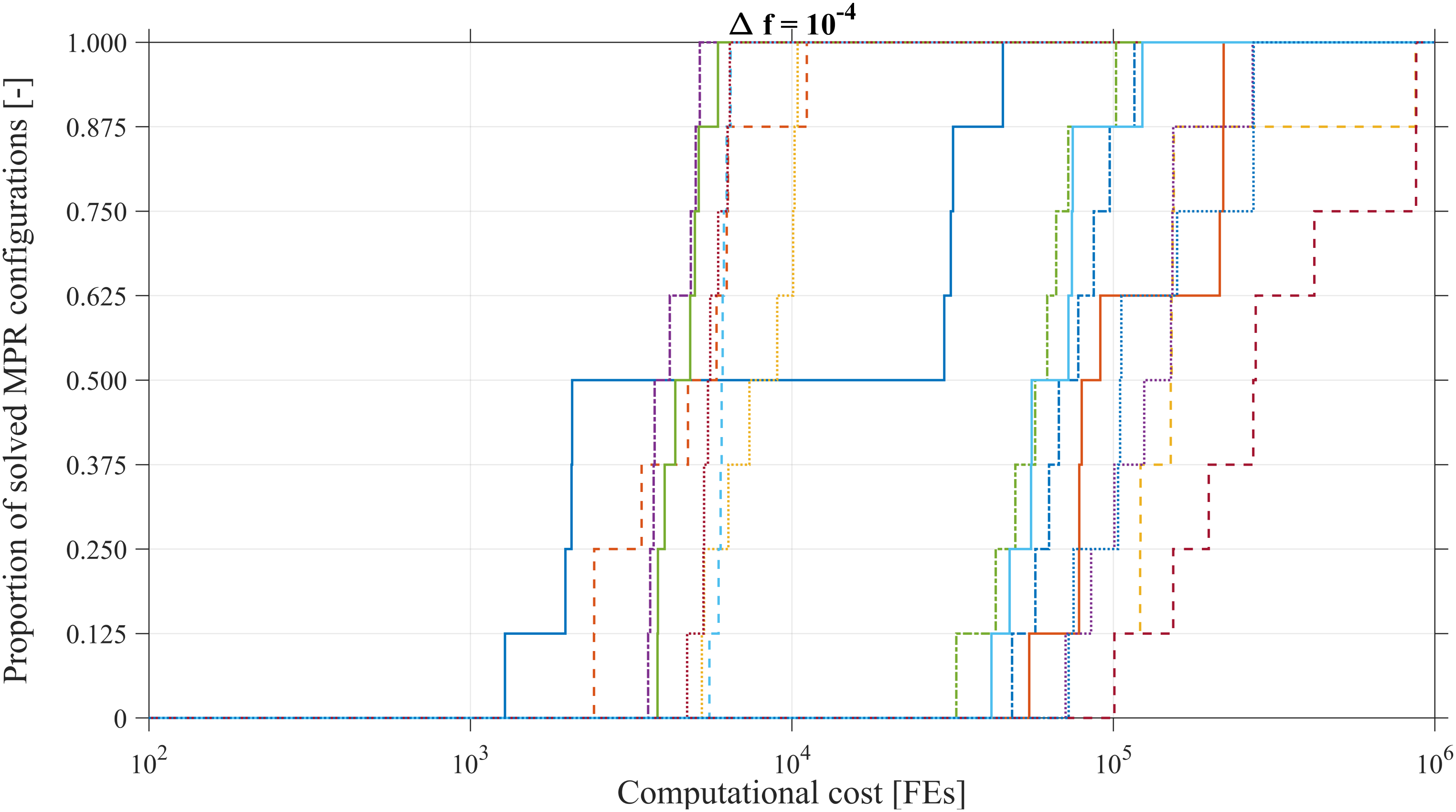}

\vspace{0.4cm}

\includegraphics[
    width=0.9\textwidth,
    keepaspectratio
]{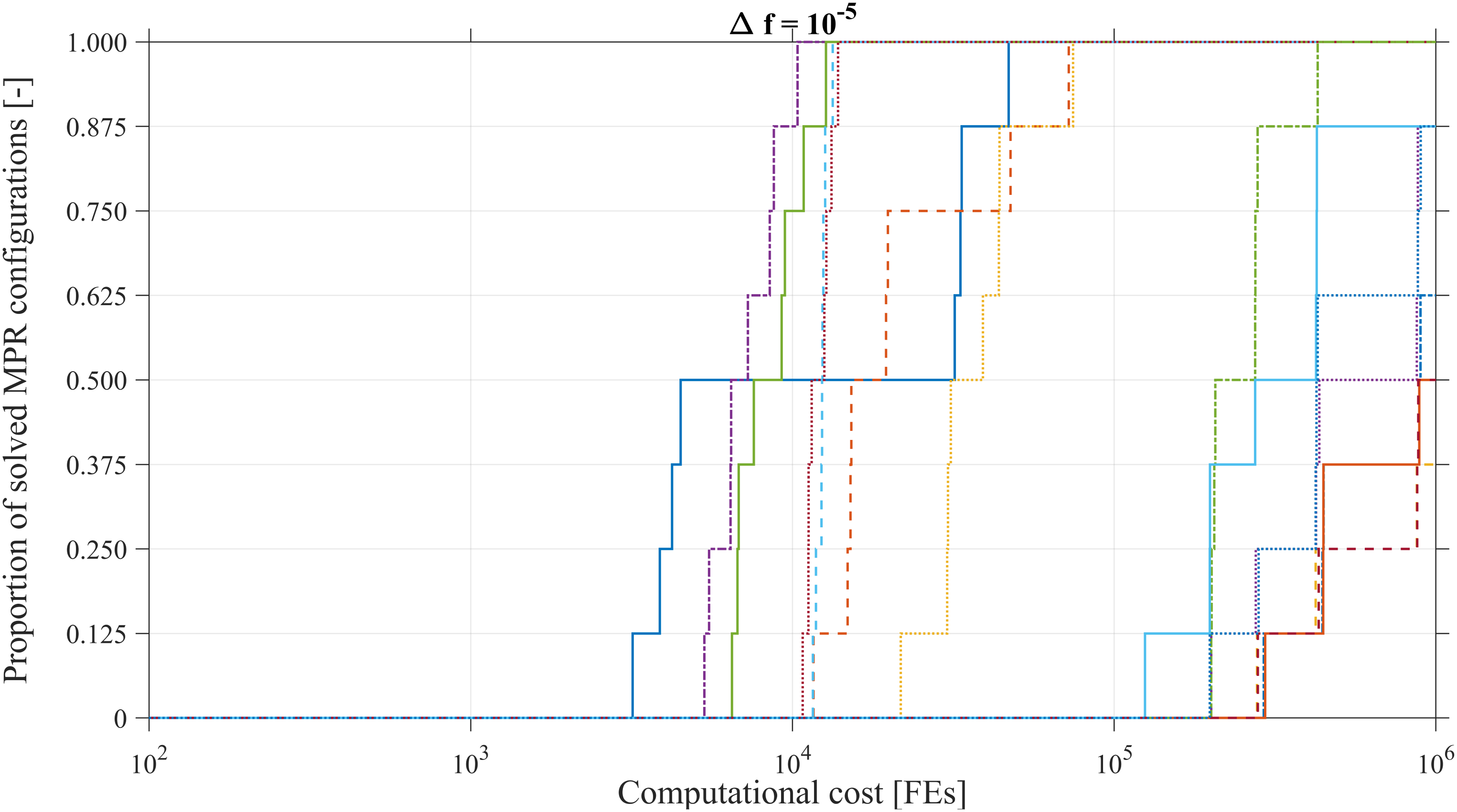}

\vspace{0.30cm}

\includegraphics[
    width=0.55\textwidth,
    keepaspectratio
]{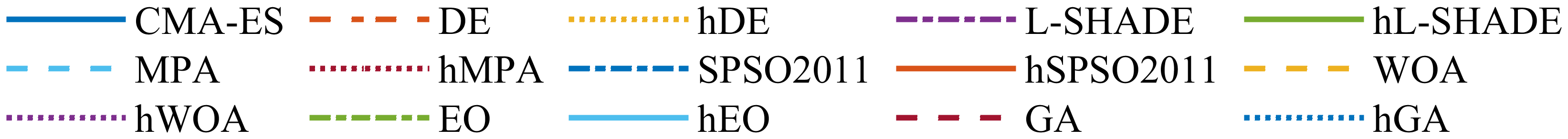}
\setcounter{figure}{2}
\caption{Fixed-target empirical runtime distributions across eight $MPR$ configurations
for $\Delta f$.}  
\label{fig:target_ert}
\end{figure}

Figure~\ref{fig:target_ert} shows the fixed-target empirical runtime distributions at the
four selected accuracy levels. Each curve corresponds to a different algorithm.

The abscissa shows the computational cost in FEs on a logarithmic scale. The
ordinate shows the proportion of the eight $MPR$ configurations for which the
estimated expected runtime,
$\widehat{E}\!\left(RT(A, C, \Delta f)\right)$, does not exceed the corresponding
computational cost threshold, $\kappa$. Each $0.125$ increase therefore corresponds
to one additional $MPR$ configuration. Configurations for which
$\widehat{E}\!\left(RT(A, C, \Delta f)\right)=+\infty$ do not increase the
corresponding curve. A curve value below one at a given computational cost threshold
indicates that, for at least one $MPR$ configuration,
$\widehat{E}\!\left(RT(A, C, \Delta f)\right)>\kappa$ or
$\widehat{E}\!\left(RT(A, C, \Delta f)\right)=+\infty$.

The four selected levels represent different target accuracies. To avoid
dependence on a single value of $\Delta f$, the analysis is extended over the
complete target grid. Following the empirical runtime aggregation of Opara and
Arabas in~\cite{KO2011}, each configuration target pair is treated as one evaluation
unit. The resulting distribution is the empirical cumulative proportion of these units 
for which the estimated expected runtime does not exceed the computational-cost threshold, $\kappa$.

\begin{figure}[!htbp]
\centering

\includegraphics[
    width=1.01\textwidth,
    keepaspectratio
]{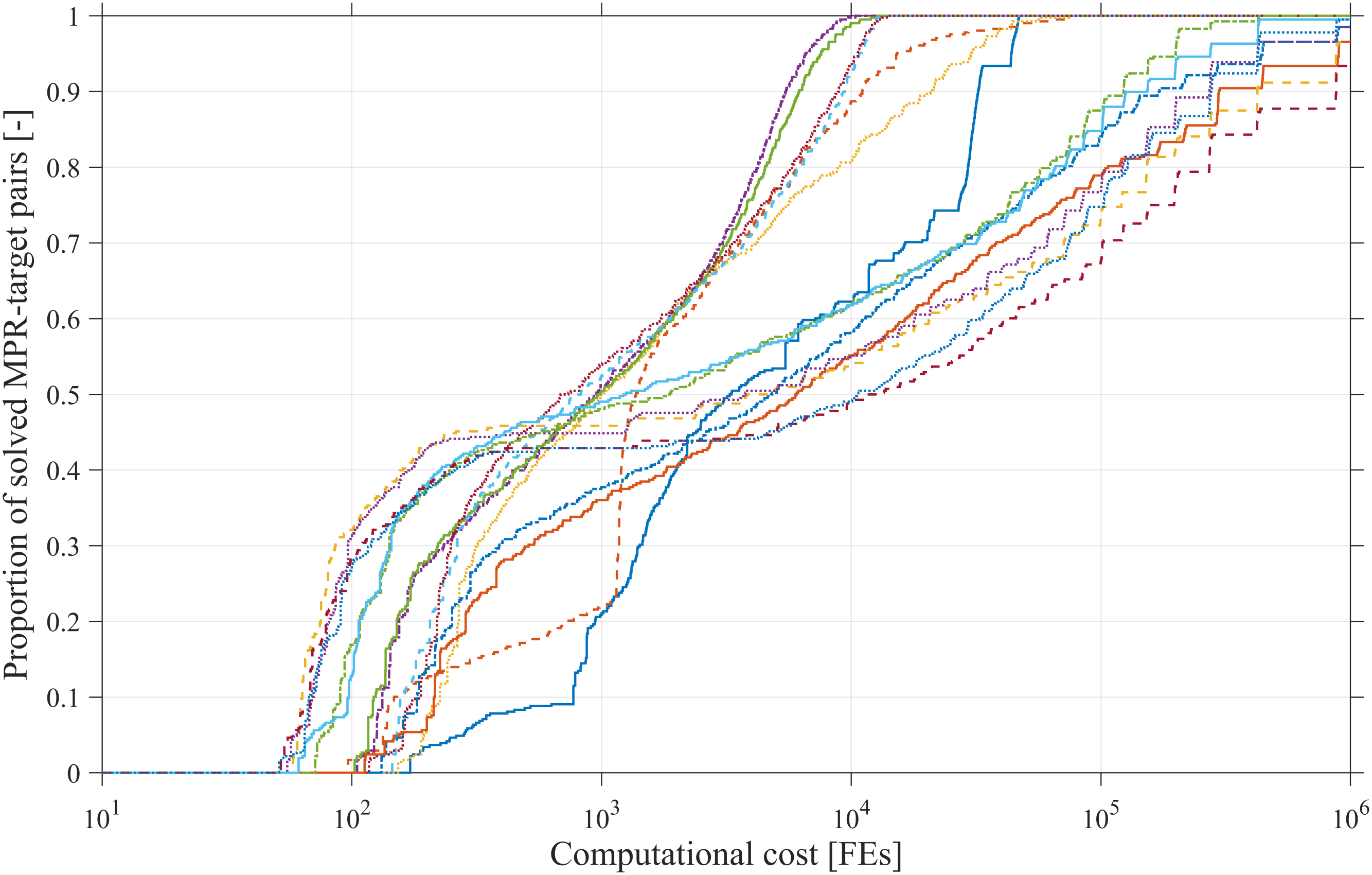}

\vspace{0.34cm}

\includegraphics[
    width=0.55\textwidth,
    keepaspectratio
]{ogLegend.png}

\caption{Aggregated empirical runtime distributions over the complete set of
$408$ $MPR$ configuration-target evaluation units.}
\label{fig:global_ert}
\end{figure}

Thus, with eight configurations and $K=51$ target levels, each algorithm is
represented by $8\cdot K=408$ units. The aggregated empirical runtime distribution is
defined as follows:
\begin{equation}  
P(A, \kappa)
=
\frac{1}{408}
\sum_{C\in\{0, 1\}^3}
\sum_{\ell=1}^{K}
\mathbf{1}
\left[
\widehat{E}\!\left(RT(A, C, \Delta f_{\ell})\right)
\leq
\kappa
\right].
\label{eq:global_ert}
\end{equation}
Here, the outer sum covers all eight $MPR$ configurations, with $A$, $C$,
$\Delta f$, $\kappa$ and $K$ defined as above.
The standard EDF representation given by Ndwandwe et al. in~\cite{LN2023}
is adopted in Eq.~(\ref{eq:global_ert}). This adapts the aggregation of Opara and Arabas in~\cite{KO2011} 
over pairs of the form $(F,\Delta f)$, where F denotes the benchmark problem and $\Delta f$ the target accuracy.
Here, $F$ is replaced by the $MPR$ configuration index, $C$, while retaining the fixed-target interpretation. 
No simulated-restart or resampling procedure is applied. The resulting distribution is shown in
Fig.~\ref{fig:global_ert}.

Figure~\ref{fig:global_ert} shows the aggregated empirical runtime distribution for all 408 configuration target units. The abscissa shows the computational cost threshold in FEs on a logarithmic scale. The ordinate shows the proportion of units for which the estimated expected runtime does not exceed the threshold. Units with an infinite expected runtime remain in the denominator and therefore do not increase the cumulative proportion.

\section{Discussion}

We use the eight $MPR$ configurations (e.g., Table \ref{tab:MPR_configs}, p.7) to conduct a controlled test of the robustness of the search mechanism. These configurations probe changes in the structure of feasibility rather than a simple scale of difficulty. As shown in Fig. \ref{fig1}, the feasible-region geometry and violation field change while the nominal objective remains fixed. The single mechanism cases $C(100)$, $C(010)$ and $C(001)$ isolate $M$, $P$ and $R$, respectively. $P$ is the most diagnostic. It causes feasibility to depend on multiple geometric realisations. $R$ reduces tolerance to inaccurate moves. $M$ alters the mechanical feasibility conditions. Therefore, we relate differences between algorithms to responses to altered violation information, movement direction and movement scale. $C(101)$, however, provides a useful contrast. Two active mechanisms without $P$ do not reproduce the responses observed for $C(110)$ and $C(011)$.

The optimisation trajectories in Fig.\ref{fig2} reinforce the non additive response. Quality progress separation, $\Delta^{+}$, is widespread. In contrast, strong violation reduction separation (denoted by $\Delta^{-}$) is concentrated mainly in $C(110)$, $C(011)$ and $C(111)$. Therefore, $MPR$ (e.g. Table \ref{tab:MPR_configs}) does not affect all methods uniformly. Specific configurations reveal the distinct limitations of their search mechanisms. The target-cost profiles in Fig.\ref{fig:target_ert} demonstrate when these discrepancies become costly as accuracy increases. Fig.\ref{fig:global_ert} examines their persistence across the full configuration accuracy range.

Configuration dependence is particularly pronounced for CMA-ES (e.g., Hansen et al. in \cite{NH2001},\cite{NH2003},\cite{NH2004}). As shown in Table \ref{tab:MPR_all_results} (pages: 13-14), full feasibility is lost in all four $P$ configurations. CMA-ES compresses samples into a single model of mean, scale, and covariance. Under $P$, different samples may be limited by different worst case geometries. Consequently, ranking merges geometrically distinct causes of success and failure. While covariance may still identify useful directions, it cannot explicitly represent intersecting feasibility conditions. This is consistent with good attainable values, but low repeatability of feasibility. The trajectories in Fig. \ref{fig2} mainly associate the effect with reduction in violations, rather than stalled quality improvement. The rising target cost in Fig.\ref{fig:target_ert} and the multistage profile in  Fig.\ref{fig:global_ert} demonstrate that this heterogeneity persists as accuracy increases. Therefore, the limitation concerns the robustness of a single model to a worst-case structure, rather than the optimisation capability itself.

A different limitation emerges for DE (e.g., Storn and Price in \cite{RS1997}). $P$ alone does not result in a complete loss of feasibility; the deficit only appears in $C(011)$ and $C(111)$ when $P$ and $R$ are both active, and it affects only a few runs. As DE's differential vector is rebuilt from the current population, it does not rely on a single global geometry model. The weaker point is late-stage step control. Fixed $F$ ($=0.5$) and $CR$ ($=0.9$) ignore success history, while the tighter margin makes small overshoots consequential. Fig.\ref{fig2} shows a milder disruption thanlate-stage for CMA-ES, although quality improvement rates still differ. Rising cost at high accuracy is evident in Figures \ref{fig:target_ert}-\ref{fig:global_ert} (pages: 22-24).

By linking selection to future moves, L-SHADE (pioneered by Tanabe and Fukunaga in \cite{RT2014}) enables accepted trials to update subsequent parameters. Each $MPR$ (e.g. Table \ref{tab:MPR_configs}) configuration therefore modifies the success statistics that drive the search. LPSR (Linear Population Size Reduction) further refines the search by altering both step dynamics and the spatial scale of the population. Across the $MPR$ family, Table \ref{tab:MPR_all_results} shows very low dispersion. Despite the clear separation of methods in terms of violation reduction under selected configurations, Fig.\ref{fig2} shows no configuration-dependent collapse, indicating preserved constraint control and quality improvement. The low target-cost sensitivity observed in Fig.\ref{fig:target_ert} persists across the wider accuracy range depicted in Fig.\ref{fig:global_ert}.

Similar robustness is achieved via a different route in MPA (proposed by Faramarzi et al. in \cite{AF2020}) across the $MPR$ configurations. Its phase dependent architecture modifies the type of movement and effective scale during the process. This avoids a single search dynamic. Full feasibility and minimal dispersion are shown in Table \ref{tab:MPR_all_results}, while stable target coverage is shown in Fig.\ref{fig2}. At high accuracy, the relatively compact cost observed in Figures \ref{fig:target_ert}-\ref{fig:global_ert} is also maintained. The common feature of MPA and L-SHADE (e.g, \cite{RT2014}) is not a specific adaptation operator. Rather, it is the progressive reduction of the effective search scale while preserving a broader search earlier in the run. Distinct architectures therefore yield similar $MPR$ robustness.

For SPSO2011 (e.g., Clerc in \cite{MC2012}) and EO (e.g. Faramarzi et al. in \cite{AFeo2020}), robustness is mainly retained at the level of feasibility. As Table \ref{tab:MPR_all_results} shows, full feasibility is preserved, and no persistent CMA-ES (e.g., Hansen et al. in \cite{NH2001},\cite{NH2003},\cite{NH2004}) like deficit appears in Fig.\ref{fig2}. Their multi-reference mechanisms reduce dependence on a single local model and promote diversity. Local success is only weakly coupled to further step-size reduction. Therefore, their limitation emerges at high accuracy, as shown in Figures \ref{fig:target_ert}-\ref{fig:global_ert} (pages: 22-24). $MPR$ does not prevent feasible solutions; rather, it increases the cost of precise refinement.

Similar patterns appear for WOA and GA, albeit through different mechanisms. WOA (e.g., Mirjalili and Lewis in \cite{SM2016}) mainly reduces movement scale with search progress rather than recent update success. GA (e.g., Deb and Kumar in \cite{KD1995} and Deb in \cite{KD2000}) rebuilds solutions through crossover and mutation, while TCSD constraints depend on nonlinear variable interactions. Therefore, good parents do not necessarily produce equally favourable offspring. Both methods retain full feasibility, as shown in Table \ref{tab:MPR_all_results}, with no persistent CMA-ES like violation deficit in Fig.\ref{fig2}. However, their weakness emerges at high accuracy, as Figs.\ref{fig:target_ert}-\ref{fig:global_ert} show, due to increased target cost. The limitation lies in the final concentration rather than in identifying the feasible region.

Hybridisation (e.g., Section \ref{sec:predicted_candidate_operator})is most effective when it addresses a specific weakness of the base method. This is most evident in hDE, where replacing weaker points reshapes the population, generating subsequent differential vectors and indirectly affecting mutation. As Table \ref{tab:MPR_all_results} shows, full feasibility in $C(011)$ and $C(111)$ is recovered, while Figures \ref{fig2}-\ref{fig:global_ert} indicate higher target reliability without materially changing the best attainable solution. Gains are smaller for hL-SHADE because success-based adaptation is already active, and the \texttt{LPSR} limits the operator's subsequent influence. In hMPA, additional information enters a mechanism that already effectively regulates movement scale. For hSPSO2011, hEO, hWOA and hGA, the operator can improve the composition of the population without altering the native point-generation rule. Thus, Figures \ref{fig2}-\ref{fig:global_ert} demonstrate that hybridisation can mitigate configuration-dependent cost or feasibility loss while the primary $MPR$ response remains tied to the base search mechanism.

\section{Conclusion}

This study shows that achieving a similar final solution does not imply that the algorithms used are similarly robust. Controlled changes to the feasibility structure reveal differences in feasibility, repeatability and the cost of precise refinement. Some of these differences remain hidden in the nominal TCSD (e.g., Section \ref{subsec:classical_TCSD}).

$MPR$ also alters the outcome of the engineering design. It alters the geometry and location of the feasible region, while the nominal objective remains unchanged. This results in different spring geometries and optimal objective values. The modification is therefore structural rather than cosmetic. It reflects the cost of satisfying stronger robustness conditions. It also distinguishes between the ability to find a good solution and the ability to reproduce it while maintaining feasibility. The contrasting behaviour reported across methods in Table \ref{tab:MPR_all_results} (pages: 13-14) supports this distinction.

The method of response depends on the configuration of the $MPR$ and the accuracy of the target. No single global ordering emerges. Some methods remain feasible but incur high refinement costs. Others only lose reliability under specific combinations of conditions. $MPR$ therefore reveals the sensitivities associated with the geometry of feasibility, movement scale and the use of search history. \texttt{Hybridisation} (e.g., Section \ref{sec:predicted_candidate_operator}) is helpful when it compensates for a specific weakness of the base mechanism. However, it does not provide universal robustness.

Changes in target accuracy further alter the relationship between the methods. The fast attainment of an approximate solution does not imply comparable efficiency at high precision. Classic endpoint statistics may therefore fail to reveal differences that are relevant in practice. Instead of collapsing them into a single ranking, landscape, feasibility, performance space and fixed-target analyses separate these effects.

The conclusions are limited to the studied TCSD (e.g., Section \ref{subsec:MPR}) formulation, the adopted $MPR$ amplitudes and the quadratic penalty with a value of $10^6$ (e.g., Equation \ref{eq:MPR_penalty}) for the parameter $\rho$. The latter was fixed throughout the present experiments. Mechanistic interpretations are supported by consistent empirical patterns rather than direct causal evidence. Open questions include penalty sensitivity, higher dimensional models, additional robustness parameters and other engineering issues. Direct measurements of population states and step dynamics would enable the proposed mechanisms to be tested more directly. \textbf{Within these limits, the $MPR$ shifts the focus of algorithm assessment from endpoint performance to robustness under controlled changes in feasibility conditions}.

\vspace{0.5cm}
\textbf{Data availability:} The results supporting the findings of this study are reported in the article. The underlying numerical datasets are available from the corresponding author upon reasonable request.

\textbf{Declaration of Competing Interest:} The authors declare that they have no known competing financial interests or personal relationships that could have appeared to influence the work reported in this paper.


\renewcommand{\refname}{
\end{document}